\documentclass{article}
\PassOptionsToPackage{square,numbers}{natbib}

\usepackage[preprint]{neurips_2026} 

\usepackage[utf8]{inputenc} 
\usepackage[T1]{fontenc}    

\usepackage{hyperref}       

\usepackage{url}            
\usepackage{booktabs}       
\usepackage{wrapfig}        
\usepackage{amsfonts}       
\usepackage{nicefrac}       
\usepackage{microtype}      
\usepackage{xcolor}         

\usepackage{natbib}
\usepackage{graphicx}
\usepackage{subcaption}
\usepackage{amsmath}
\usepackage{bm}  
\usepackage{adjustbox}
\usepackage{tikz}
\usetikzlibrary{arrows.meta,positioning,calc}
\usepackage{amssymb}

\usepackage{float}

\title{CrystalMO-TuRBO: Multi-Objective Trust-Region Bayesian Optimization for High-precision Joint Crystal Structure Refinement}

\author{%
  Joseph Agada\thanks{Corresponding Author.} \\
  Bredesen Center for Interdisciplinary Research\\
  University of Tennessee\\
  Knoxville, TN 37996 \\
  \texttt{joe88data1@gmail.com} \\
  \And
  Yishu Wang \\
  Department of Materials Science and Engineering \\
  University of Tennessee Knoxville, TN 37996 \\
  \texttt{wangyishu@utk.edu} \\
  \AND
  Arpan Biswas\footnotemark[1] \\
  University of Tennessee - Oak Ridge Innovation Institute \\
  University of Tennessee Knoxville, TN 37996 \\
  \texttt{abiswas@utk.edu} \\
}

\begin{document}

\maketitle

\begin{abstract}
Crystal structure refinement is a fundamental inverse problem in materials characterization, where structural parameters are optimized to reproduce experimental diffraction data. Conventional approaches, such as least-squares and likelihood-based optimization, rely on local search and often struggle with non-convex, noisy, and highly correlated parameter landscapes, particularly when integrating multiple diffraction modalities. Joint refinement of X-ray and neutron data is especially challenging due to their complementary but competing sensitivities, which are typically combined through scalarized objectives requiring manual weighting and leading to suboptimal solutions. We propose \textbf{CrystalMO-TuRBO}, a multi-objective trust-region Bayesian optimization architecture for joint crystal structure refinement. The method models X-ray and neutron discrepancies as separate objectives and transforms the problem into a normalized maximization setting. A two-phase optimization strategy is introduced: Phase~1 performs global exploration using parallel trust-region Bayesian optimization across multiple scalarizations to identify promising regions of the parameter space, while Phase~2 conducts localized refinement within a shrinking region to achieve high-precision solutions. This design explicitly separates global search from fine-grained optimization, addressing the unique accuracy requirements of refinement tasks. We evaluate the proposed method on experimentally collected X-ray and neutron diffraction data from single-crystal Ho$_2$Ti$_2$O$_7$. Results demonstrate improved convergence, robustness, and parameter precision compared to classical refinement methods and Bayesian optimization baselines on refinement of a single-crystal pyrochlore material system.
\end{abstract}

\section{Introduction}

Many scientific discovery tasks can be formulated as \emph{inverse problems}, where latent physical parameters must be inferred from indirect, noisy, and computationally expensive observations generated by a forward model. Crystal structure refinement is a canonical example, in which structural, defect, and instrumental parameters are optimized so that simulated diffraction patterns closely reproduce experimental measurements \cite{IUCrRefinement}. Because material properties are governed by atomic arrangement, occupancies, lattice distortions, and magnetic ordering, accurate refinement is fundamental to materials characterization and structure--property analysis \cite{Egami2012,Martin2006,Billinge2019}. Diffraction techniques such as X-ray diffraction (XRD) and neutron diffraction (ND) provide complementary but indirect information about crystal structures, making refinement a challenging inverse problem involving expensive objective evaluations, strong parameter correlations, and highly multimodal optimization landscapes.

From a machine learning perspective, crystal structure refinement presents several characteristics that make it well suited to Bayesian optimization. The objective landscape is highly non-convex, exhibits multiple locally optimal refinement basins, and each objective evaluation requires solving a computationally expensive physics-based forward model. Conventional refinement methods, including nonlinear least-squares (Rietveld refinement), maximum likelihood estimation, Bayesian inference, and simulated annealing, formulate refinement as a \emph{single-objective} optimization problem \cite{Rietveld1969,Young1993,Bricogne1997,Sivia2006,Kirkpatrick1983}. Although these approaches remain the dominant workflows in crystallography, they are sensitive to initialization, susceptible to local minima, and often struggle to achieve the extremely high precision required for complex material systems \cite{Toby2006}. Recently, Agada \emph{et al.}~\cite{Agada2025BORefinement} demonstrated that Bayesian optimization can improve crystal structure refinement by replacing local optimization with a global surrogate-based search. However, their formulation remained strictly single-objective with refinement conducted using only one (XRD) modality.

A fundamental limitation of single-objective refinement becomes apparent when complementary experimental measurements are available. XRD primarily probes electron density, whereas ND is sensitive to nuclear positions and magnetic structure \cite{Young1993,Egami2012}. Consequently, independently optimizing the two modalities generally produces different optimal parameter sets for the same crystal because each modality emphasizes different physical characteristics. Conventional joint refinement addresses this by minimizing a weighted sum of the individual residuals \cite{Afonine2010,Deutsch2012}. However, this scalarization requires user-defined weighting factors and obscures the trade-offs between competing experimental objectives, potentially biasing the refinement toward one modality \cite{Miettinen1999}. This motivates reformulating joint refinement as a \emph{multi-objective Bayesian optimization} (MOBO) problem, where XRD and ND objectives are optimized simultaneously without collapsing them into a single weighted objective.

In this work, we propose \textbf{CrystalMO-TuRBO} (CMT), a multi-objective trust-region Bayesian optimization framework for joint crystal structure refinement from complementary diffraction measurements. The framework models the XRD and ND refinement objectives using Gaussian process surrogates \cite{Rasmussen2006,Snoek2012} and combines the sample efficiency of Bayesian optimization with the robustness of trust-region search. Unlike conventional global BO, the proposed framework employs multiple trust regions to efficiently explore multiple promising refinement basins before transitioning to localized high-precision optimization. Although trust-region Bayesian optimization was originally introduced for high-dimensional optimization, its use here is motivated by the highly multimodal and heterogeneous refinement landscape rather than by dimensionality alone \cite{Eriksson2019TuRBO,Paria2020}. The proposed two-phase design therefore first identifies promising regions of the parameter space and subsequently performs fine-grained refinement by shrinking the search region while retaining all optimization parameters.

Unlike conventional optimization benchmarks, crystal structure refinement requires extremely high parameter accuracy rather than approximate optimality. The proposed framework addresses this requirement by combining global exploration with localized refinement while maintaining the multi-objective structure throughout the optimization process. As a result, the method identifies a single physically consistent parameter set that jointly minimizes the XRD and ND objectives without relying on heuristic weighting during optimization.

We evaluate the proposed framework using experimentally collected XRD and ND data from a single pyrochlore $Ho_{2}Ti_{2}O_{7}$ crystal. Experimental results demonstrate that CMT consistently outperforms conventional least-squares joint refinement, representative multi-objective Bayesian optimization methods, including ParEGO \cite{Knowles2006ParEGO}, qEHVI \cite{Daulton2020qEHVI}, and MORBO \cite{Daulton2022MORBO}. Beyond the pyrochlore system considered here, the proposed framework provides a general optimization strategy for expensive multi-objective inverse problems involving complementary experimental modalities.

\section{Related Work}
\label{sec:related_work}

Crystal structure refinement is traditionally formulated as an optimization problem in which structural and instrumental parameters are adjusted to minimize the discrepancy between experimental and simulated diffraction patterns. The dominant approach is nonlinear least-squares optimization, most notably Rietveld refinement \cite{Rietveld1969,Young1993}, which remains the de facto standard in crystallography due to its effectiveness and implementation in widely used refinement software such as GSAS-II and FullProf \cite{Toby2006}. Alternative optimization strategies, including maximum likelihood estimation (MLE), Bayesian inference, simulated annealing, and genetic algorithms \cite{Bricogne1997,Pannu1998,Sivia2006,Fancher2016,Kirkpatrick1983}, largely optimize the same underlying scalarized refinement objective using different optimization or statistical formulations. Although these methods improve robustness in certain settings, they remain susceptible to local minima, strong parameter correlations, and the highly multimodal landscapes encountered in complex crystal structure refinement.

Joint refinement combines complementary experimental modalities, most commonly X-ray diffraction (XRD) and neutron diffraction (ND), to improve parameter identifiability and refinement accuracy \cite{Squires2012,Afonine2010,Deutsch2012}. Existing joint refinement workflows, however, almost exclusively formulate the problem as a weighted single-objective optimization by combining multiple datasets into a scalar objective function. Consequently, the refinement depends on user-specified weighting factors that determine the relative influence of each dataset and may bias the solution toward one experimental modality. Although frameworks such as PHENIX \cite{Afonine2010} and subsequent extensions \cite{Liebschner2023} improve the underlying physical models, they do not fundamentally change the optimization paradigm, which remains scalarized least-squares or likelihood-based refinement.

Bayesian optimization (BO) has recently emerged as an attractive alternative for expensive scientific optimization problems because it replaces repeated evaluations of computationally expensive objective functions with surrogate-guided sequential search \cite{Shahriari2016,Frazier2018, BISWAS2025853, Harris2024Autonomous}. Agada \emph{et al.}~\cite{Agada2025BORefinement} first demonstrated the feasibility of Bayesian optimization for crystal structure refinement by formulating refinement as a \emph{single-objective} BO problem. While that work showed improved optimization over conventional least-squares refinement, each diffraction modality was still optimized independently, leaving unresolved the challenge of obtaining a single physically consistent parameter set from complementary experimental measurements.

Multi-objective Bayesian optimization (MOBO) addresses this limitation by explicitly modeling multiple competing objectives and searching directly for Pareto-optimal solutions rather than requiring predefined scalarization weights \cite{Knowles2006ParEGO,Paria2020,Biswas2021MOBO,Biswas2023AMB,Biswas2022TchebycheffMOBO, Biswas2021MOBO, moboweightedtchebychev}. Representative approaches include ParEGO \cite{Knowles2006ParEGO}, which employs scalarization, qEHVI \cite{Daulton2020qEHVI}, which maximizes expected hypervolume improvement, and MORBO \cite{Daulton2022MORBO}, which extends trust-region Bayesian optimization to the multi-objective setting. However, these methods have not been applied to crystal structure refinement, nor do they explicitly address the high-precision refinement requirements of scientific inverse problems.

Our work bridges these gaps by reformulating joint crystal structure refinement as a true multi-objective Bayesian optimization problem. Unlike existing joint refinement workflows, the proposed framework simultaneously optimizes complementary XRD and ND objectives. Furthermore, we develop a novel two-phase TuRBO-m-based optimization framework that combines robust global exploration with high-precision local refinement, making it particularly suitable for expensive and highly multimodal refinement landscapes.

\section{Methods}
\label{sec:method}

This section first introduces the crystal structure refinement problem from both the materials science and machine learning perspectives. We begin by describing the experimental X-ray diffraction (XRD) and neutron diffraction (ND) datasets, the refinement parameters, and their physical significance. We then formulate joint crystal structure refinement as a multi-objective Bayesian optimization problem by defining the XRD and ND objective functions together with the optimization parameter space. Finally, we present the proposed \textit{CMT} framework, including its two-phase trust-region Bayesian optimization strategy for jointly refining complementary diffraction measurements.

\subsection{Scattering Experiment and Data Generation}
\label{sec:experiment2}

To evaluate the proposed framework, X-ray diffraction (XRD) and neutron diffraction (ND) experiments were performed on a single-crystal $Ho_{2}Ti_{2}O_{7}$ sample to obtain the experimental diffraction datasets used throughout this work. Each experiment produces a set of indexed Bragg reflections together with their corresponding measured diffraction intensities, which serve as the experimental observations for crystal structure refinement.

The X-ray diffraction experiment follows the procedure described in \cite{Agada2025BORefinement} and is summarized in Appendix~\ref{sec:experiment}. Here, we describe the \textbf{single-crystal neutron diffraction} experiment. Single-crystal neutron diffraction measurements were performed on a $2 \times 2 \times 2$ mm$^3$ floating-zone-grown Ho$_2$Ti$_2$O$_7$ crystal at 110 K using TOPAZ, the BL-12 time-of-flight single-crystal diffractometer at the Spallation Neutron Source, Oak Ridge National Laboratory. TOPAZ employs wavelength-resolved Laue diffraction with a broad neutron wavelength band and an array of time-of-flight area detectors, enabling efficient three-dimensional reciprocal-space mapping of both nuclear and magnetic Bragg scattering. The data were reduced using the standard TOPAZ single-crystal diffraction workflow with detectors calibrated at the beginning of the experimental cycle. Events were loaded over a time-of-flight range of 1500--16600~$\mu$s and scaled by proton charge. A UB matrix was used to index reflections in the cubic $F$-centered cell. Predicted reflections were integrated over wavelengths of 0.4--3.5~\AA{} and $d$-spacings of 0.5--12.0~\AA{} using ellipsoidal peak integration with adaptive $Q$-dependent background estimation. Final intensity normalization employed the TOPAZ spectrum calibration together with a polyhedral absorption correction and $m\bar{3}m$ point-group symmetry.

\subsection{BO Formulation, Parameters and Objectives}
\label{sec:BO_form}

Let the experimentally measured diffraction intensities obtained from either the X-ray or neutron diffraction experiment described in Section~\ref{sec:experiment2} be denoted by
$
I_{\mathrm{obs}}=\{I_{\mathrm{obs},i}\}_{i=1}^{N},
$
where $N$ is the number of indexed Bragg reflections. Each reflection corresponds to a specific set of crystal planes, and its measured intensity encodes information about the underlying crystal structure. Crystal structure refinement seeks the model parameters that best reproduce these experimentally observed intensities. For a given reflection $i$, the corresponding calculated intensity is
$
I_{\mathrm{calc},i}(\theta,\vartheta)
=
kL_iP_iT_i
y_i(\vartheta)
|F_{c,i}(\theta)|^2,
$ where $k$ is an overall scale factor, $L_i$, $P_i$, and $T_i$ are fixed Lorentz, polarization, and absorption/transmission correction factors determined by the experimental geometry \citep{Becker1974ExtinctionI,Becker1974ExtinctionII}, $|F_{c,i}(\theta)|^2$ is the squared structure factor determined by the crystal structure, and $y_i(\vartheta)$ is the extinction correction accounting for multiple scattering. The optimization variables consist of two groups. The structural parameters
$
\theta=
\{
\mathrm{Occ}{\mathrm{Ti}},
U{\mathrm{Ti}},
\mathrm{Occ}{\mathrm{Ho}},
U{\mathrm{Ho}},
x{\mathrm{O1}},
\mathrm{Occ}{\mathrm{O2}},
U{\mathrm{O1}},
U{\mathrm{O2}}
\},
$ describe the crystal structure. Here, $\mathrm{Occ}$ denotes atomic occupancy, $U$ denotes the atomic displacement parameter describing thermal vibration about equilibrium positions, and $x_{\mathrm{O1}}$ is the fractional coordinate of the O1 atom within the unit cell. The extinction parameters $
\vartheta=\{E_p,E_s,E_g\},
$ model primary, secondary, and Gaussian extinction effects that influence neutron diffraction intensities \citep{Becker1974ExtinctionI,Becker1974ExtinctionII}. Unlike the structural parameters, these describe instrument- and sample-dependent scattering effects rather than the crystal structure itself. The refinement problem is therefore formulated as $
(\theta^*,\vartheta^*,k^*)
=
\arg\min_{\theta,\vartheta,k}
f(\theta,\vartheta,k),
$ where the objective measures the discrepancy between the observed and calculated diffraction intensities. The objective function $f(\theta,\vartheta,k)$ is defined, for XRD and ND as:



\begin{equation}
f_m=
\frac{1}{\alpha_m N}
\sum_{i=1}^{N}
\left(
I_{\mathrm{obs},i}^{(m)}
-
I_{\mathrm{calc},i}^{(m)}(\theta,\vartheta,k_m)
\right)^2,
\quad
m\in\{\mathrm{XRD},\mathrm{ND}\},
\label{eq:chisq}
\end{equation}

where the weighting factors ($\alpha_{XRD} = 1000$ and $\alpha_{ND} = 1$) place the XRD and ND objectives on comparable numerical scales because X-ray diffraction intensities are typically several orders of magnitude larger than neutron diffraction intensities. Although both modalities share the same structural parameters, their forward models differ in two important ways. First, the structure factor is computed differently because X-rays scatter from electron density whereas neutrons scatter from atomic nuclei; the corresponding formulations are given in Equations~\ref{eq:Fcal_xrd} and~\ref{eq:Fcalc_nd}. Second, extinction is modeled only for neutron diffraction. Accordingly, $y_i(\vartheta)=1$ for XRD, and the extinction parameters are optimized only when neutron diffraction data are included. To ensure physically meaningful solutions, the optimization variables are constrained within crystallographically plausible bounds. The occupancies of Ti, Ho, and O2 are restricted to $(0.8,1.2)$, atomic displacement parameters to $(0,0.1)$, and the O1 fractional coordinate to $(0.35,0.45)$. Candidate solutions outside these bounds are excluded from the search space.

\subsection{Objectives, Design Variables, and Objective Transformation}

The optimization variables are collected into a design vector
$
\mathbf{z}\in\mathcal{B}\subset\mathbb{R}^{d},
$
comprising the structural, extinction, and scale parameters introduced in Section~\ref{sec:BO_form}. For each candidate solution $\mathbf{z}$, the forward crystallographic model evaluates the XRD and ND objective functions, $f_{\mathrm{XRD}}$ and $f_{\mathrm{ND}}$, defined in Eq.~\ref{eq:chisq}. The design variables are first normalized to $[0,1]^d$, while the objective values are standardized and sign-inverted so that the original minimization problem becomes a two-objective maximization problem compatible with Bayesian optimization acquisition functions. This normalization also improves Gaussian process conditioning and balances the numerical scales of the two diffraction objectives. The complete normalization procedure is provided in Appendix~\ref{sec:normalization}.

\subsection{Phase 1: Parallel Multi-Objective Trust-Region Search}
\label{subsec:phase1}

Phase~1 extends TuRBO~\cite{Eriksson2019TuRBO} to joint XRD--ND refinement by decomposing the two-objective problem into $K$ parallel scalarized trust-region optimization problems. Each trust region optimizes a different convex combination of the transformed objectives,
$
S_k(\mathbf{Y})=\lambda_kY_1+(1-\lambda_k)Y_2,
$
where $\lambda_k\in(0,1)$ controls the preference between XRD and ND refinement. In our implementation, four trust regions are used with $\lambda=\{0.15,0.45,0.65,0.85\}$.
Each scalarization maintains an independent TuRBO-1 state, Gaussian process surrogate, and adaptive trust region following the standard TuRBO framework~\cite{Eriksson2019TuRBO}. Candidate solutions are selected by maximizing the $q$-Expected Improvement acquisition function within each trust region, evaluated on both diffraction objectives, and appended to a shared global archive. Phase~1 begins with Latin Hypercube Sampling~\cite{McKay1979LHS} and continues until the prescribed exploration budget is exhausted. Figure~\ref{fig:Fig7-CrystalMO-TuRBO} summarizes the complete workflow.

\subsection{Phase 2: Local Trust-Region Refinement}
\label{sec:phase_two}

After Phase~1 identifies promising regions of the search space, Phase~2 performs local refinement around the current best solution using a single balanced scalarization,
$
S^{(2)}(\mathbf{Y})
=
\lambda^{(2)}Y_1+
(1-\lambda^{(2)})Y_2,
$
where $\lambda^{(2)}=0.5$. A Gaussian process surrogate and $q$-Expected Improvement acquisition function are then optimized within a fixed axis-aligned local trust region centered on the current best solution. This stage improves refinement accuracy while preserving the physically constrained search space defined in Section~\ref{sec:BO_form}. Additional implementation details, including trust-region construction and stopping criteria, are provided in Appendix~\ref{sec:phase_two_appendix}. For reporting a single refinement result, we select the non-dominated solution that minimizes the normalized distance to the utopia point (equivalently, the smallest normalized $f_{\mathrm{XRD}}+f_{\mathrm{ND}}$ in our experiments). The complete Pareto filtering and compromise-selection procedure is provided in Appendix~\ref{sec:utopia-compromise}.

 \section{Experiment and Result}
 \label{experiment_and_results}

In this work, we evaluate the proposed \textit{CrystalMO-TuRBO} (CMT), a multi-objective Bayesian optimization framework for joint crystal structure refinement using complementary X-ray diffraction (XRD) and neutron diffraction (ND) data. A single-crystal $Ho_{2}Ti_{2}O_{7}$ sample was synthesized, and XRD and ND experiments were performed to obtain the diffraction datasets used throughout this study. All analyses were conducted on a Windows 11 personal computer equipped with an Intel Core i7-12700H processor, 16 GB RAM, and a 500 GB SSD. The implementation was developed in Python using GSAS-II for crystallographic refinement and GPyTorch and BoTorch for Gaussian process modeling and Bayesian optimization, respectively. The proposed CMT required approximately 2 hours per run, compared with approximately 2 hours 20 minutes for MORBO and about 6 hours for each of qEHVI and ParEGO. For comparison, we implemented the conventional least-squares joint refinement workflow, which is the community standard for crystallographic refinement, together with representative multi-objective Bayesian optimization baselines, including qEHVI \cite{Daulton2020qEHVI}, MORBO \cite{Daulton2022MORBO}, and ParEGO \cite{Knowles2006ParEGO}. All Bayesian optimization methods were executed using an identical evaluation budget of 400 function evaluations. The XRD and ND objective functions are those defined in Eq.~\ref{eq:chisq}. Results for joint refinement with primary and secondary extinction correction are summarized in Table~\ref{table:pe-se-combined}. Each subtable reports the conventional least-squares baseline, the proposed CMT under different local bounding-box widths (1\%, 2\%, 5\%, and 10\%), and the representative MOBO baselines. For CMT, results are reported for both Phase~1 alone and the complete Phase~1 + Phase~2 framework to isolate the contribution of the proposed local refinement stage. To evaluate robustness, we performed average of 12 independent runs of the least-squares and all multi-objective Bayesian optimization methods using different random seeds. The repeated-run statistics are reported in Tables~\ref{tab:table1} and~\ref{tab:table2}. Pairwise comparisons between competing methods were performed using the Mann--Whitney U test, and the observed improvements in the relevant objective values were confirmed to be statistically significant at 0.05 level of significance. Figures~\ref{fig:Fig13-pareto}, ~\ref{fig:Fig12-conv_curve}, ~\ref{fig:Fig11-gp_map} and ~\ref{fig:Fig10-params_conv.png} compare the Pareto fronts, optimization convergence, Gaussian process mean maps, and parameter convergence obtained by the competing methods, respectively. Figures~\ref{fig:multi-run_comparison_all} and~\ref{fig:multi-run_comparison} summarize the repeated-run analyses, illustrating the superior performance and robustness of the proposed method against MOBO baselines, and the contribution of the Phase~2 local refinement strategy to the CMT design, respectively. Additional visualizations, including enlarged versions of the comparison figures, scatter plots of observed versus calculated diffraction intensities, Gaussian process maps, and single-objective Bayesian optimization results, are provided in the Appendix.

\subsection{Benchmark and Comparison}

For a fair comparison, all multi-objective BO methods were allocated a budget of 400 function evaluations, while each single-objective BO method was allocated 200 evaluations. This budget reflects the fact that separate single-objective optimizations are required for XRD and ND, whereas a single 400-iteration MOBO run jointly optimizes both objectives. Among the single-objective BO methods, TuRBO-1 \cite{Eriksson2019TuRBO} achieved the best performance, obtaining $f_{\mathrm{XRD}}=8.82$, $f_{\mathrm{ND}}=11.02$, and $f_{\mathrm{ND}}=10.04$ for XRD refinement, ND refinement with primary extinction correction, and ND refinement with secondary extinction correction, respectively. Tables~\ref{table:pe-se-combined}, \ref{table4}, \ref{table5}, and \ref{table6} summarize the refinement results. Compared with the crystallographic community standard least-squares joint refinement, the proposed \emph{CrystalMO-TuRBO} (CMT) substantially reduces both XRD and ND mismatch while identifying a single compromise solution. For example, under primary extinction correction, least-squares refinement achieved $f_{\mathrm{XRD}}=28.50$ and $f_{\mathrm{ND}}=28.74$, whereas CMT reduced these to $11.46$ and $11.67$, respectively. Under secondary extinction correction, CMT achieved $f_{\mathrm{XRD}}=10.36$ and $f_{\mathrm{ND}}=10.46$, compared with $14.69$ and $41.14$ for least-squares refinement. Notably, the joint refinement performance of CMT approaches that of the corresponding single-objective BO methods while simultaneously optimizing both diffraction modalities. CMT also consistently outperforms the representative MOBO baselines, including qEHVI, ParEGO, and MORBO.

\begin{figure}[H]
    \centering
    \includegraphics[width=0.9\linewidth]{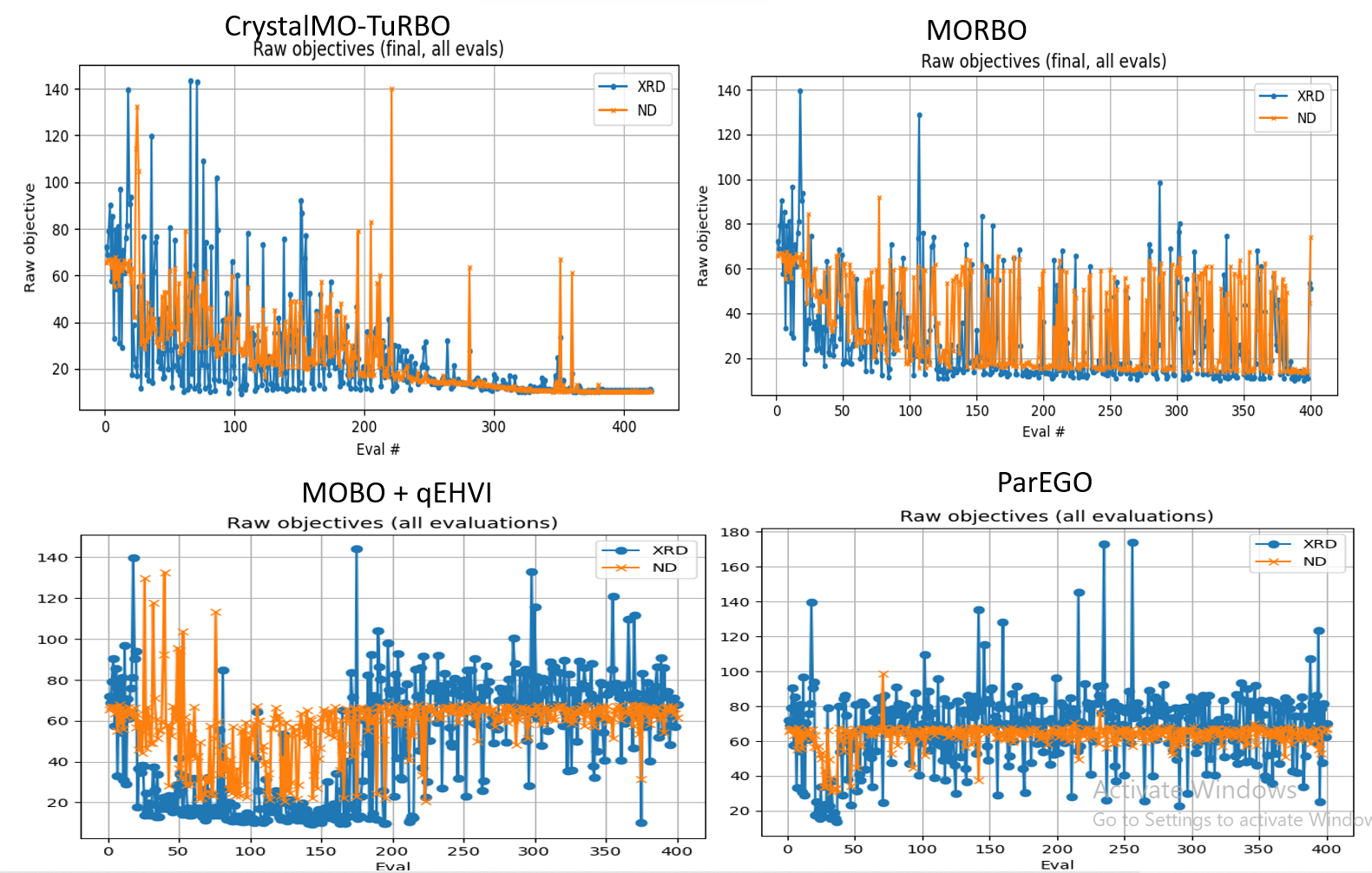}
    \caption{Figure showing the convergence curves for \emph{CrystalMO-TuRBO} and the MOBO baselines. Blue and yellow bars are for XRD and ND mismatch, respectively. As observed, \emph{CrystalMO-TuRBO} converges steadily from iteration 1 to approximately 200, after which it progressively refines the solution in a fine-grained manner until the full budget of 400 evaluations is reached. This behavior, especially after 200 iterations (phase 1), clearly distinguishes it from the other BO baselines.}
    \label{fig:Fig12-conv_curve}
\end{figure}

\begin{figure}[t]
    \centering
    \includegraphics[width=\linewidth]{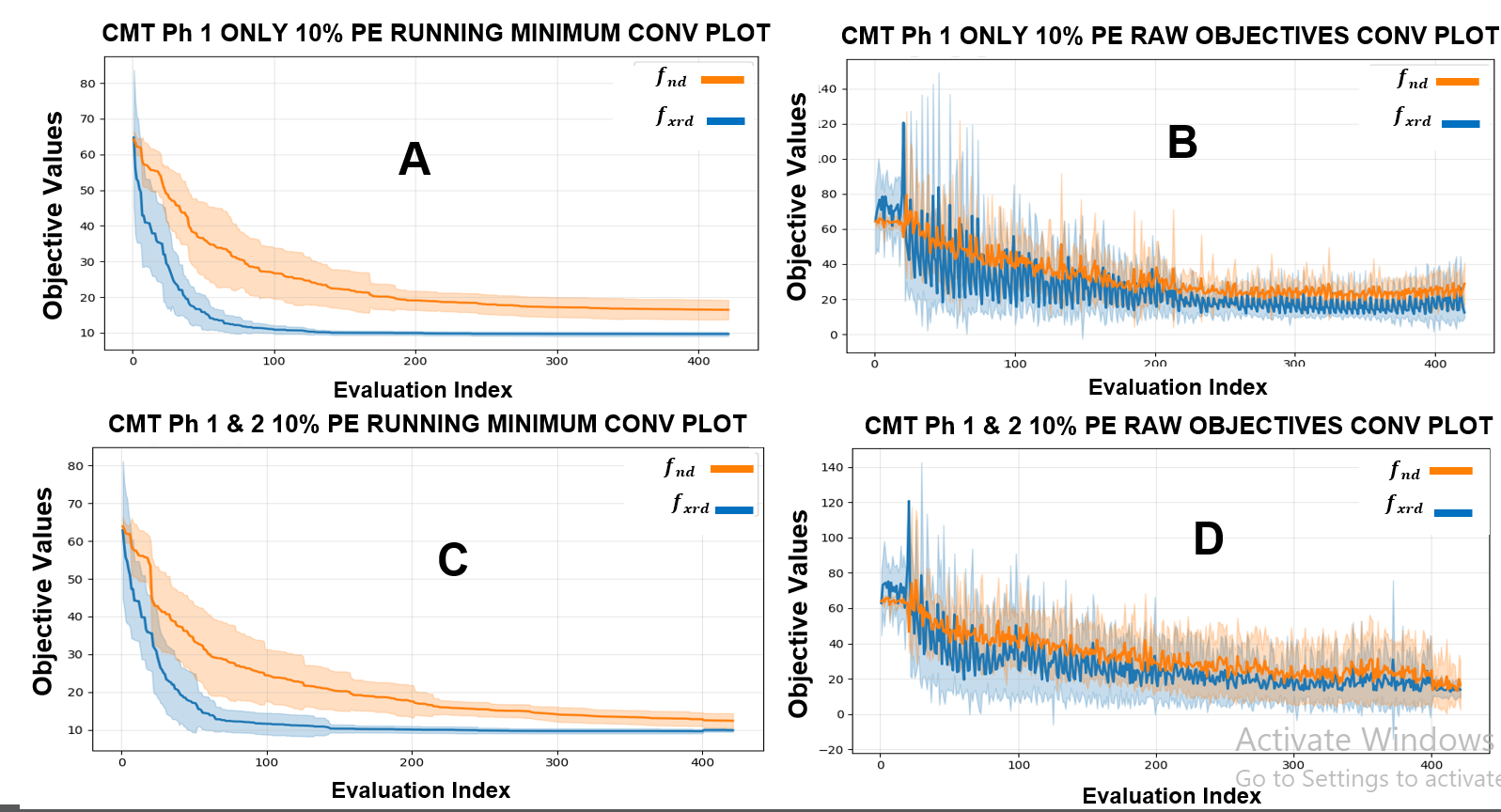}
    \caption{Figure showing convergence curve showing running minimum (Column 1) and raw objective values (Column 2). The values plotted are means and standard deviation of objective values. As visually depicted, Plots C and D, which are convergence curves for Ph 1 + Ph 2 implementation show better convergence then Plots A and B which are convergence curves for Ph 1 implementation alone}
    \label{fig:multi-run_comparison}
    \vspace{-2mm}
\end{figure}

\begin{figure}[H]
    \centering
    \includegraphics[width=0.9\linewidth]{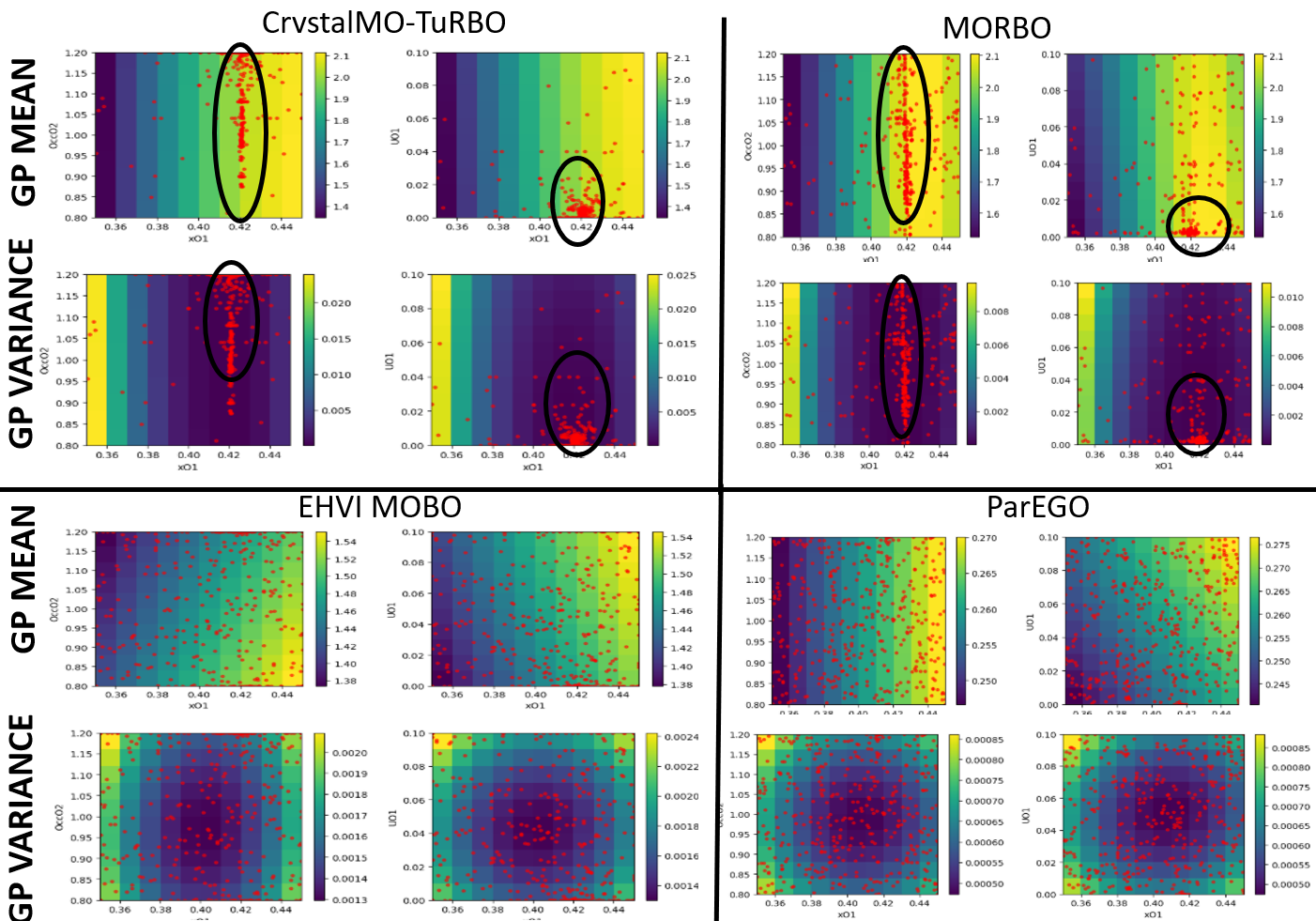}
    \caption{Figure showing GP Mean and Variance plot for $XO1$ vs $OccO2$ and $UO1$ vs $XO1$ for the proposed \emph{CrystalMO-TuRBO} and the other baselines.While \emph{CrystalMO-TuRBO} and MORBO were able to identify and focus on the promising region of the parameter space, qEHVI + MOBO and ParEGO failed in this area}
    \label{fig:Fig11-gp_map}
\end{figure}

\begin{figure}[H]
    \centering
    \includegraphics[width=0.9\linewidth]{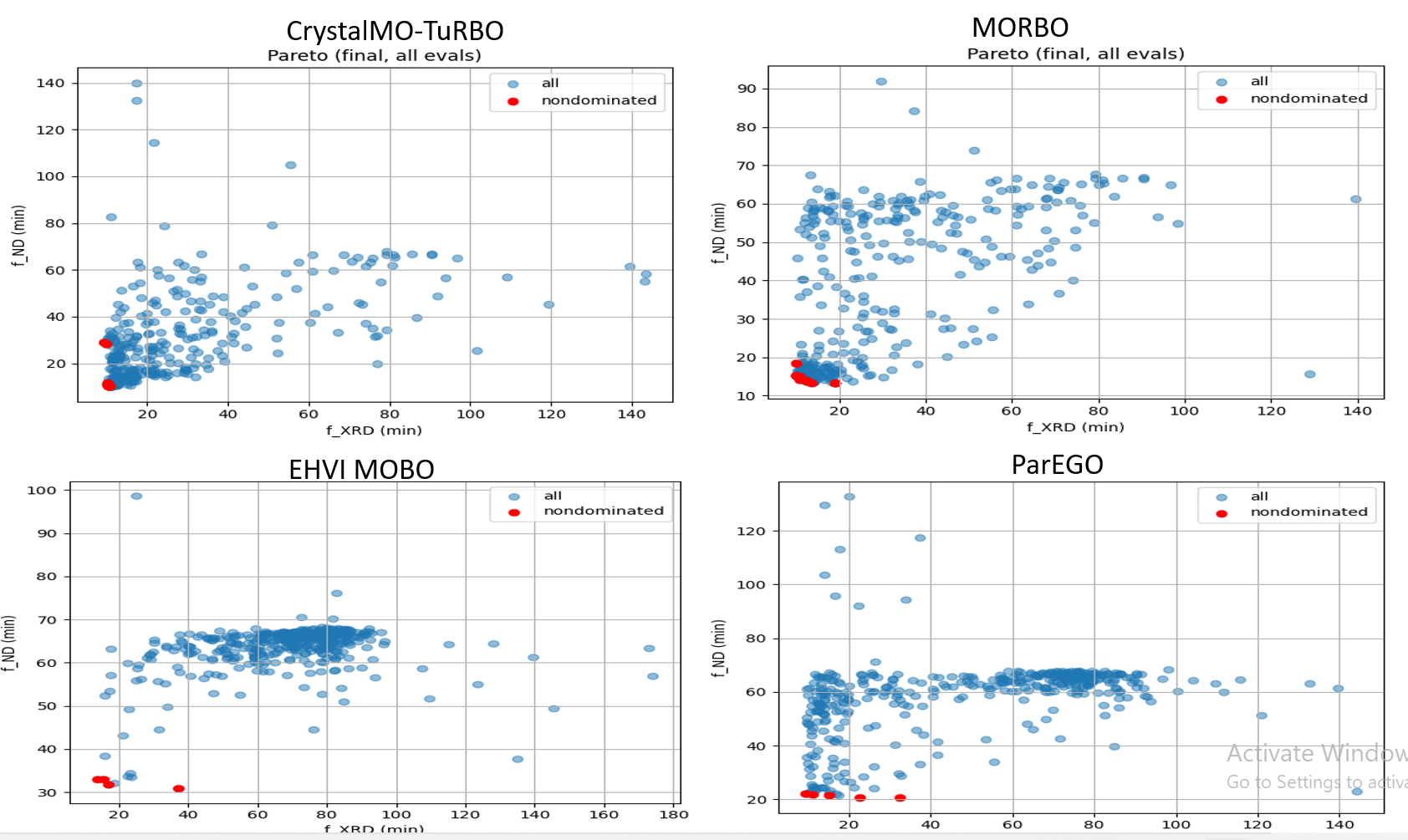}
    \caption{Figure showing the (Pareto) scatter plots for \emph{CrystalMO-TuRBO} and the baselines. As seen, while \emph{CrystalMO-TuRBO} and MORBO steadily reduce the mismatch for both objectives, $f_{ND}$ for qEHVI MOBOna d ParEGO appear to be mostly constant, while only $f_{XRD}$ reduced with iteerations. Also, as seen, the nondominant points (in red) for the proposed \emph{CrystalMO-TuRBO} are more concentrated in one location, indicating more robustness and precision.}
    \label{fig:Fig13-pareto}
\end{figure}

\begin{figure}[H]
    \centering
    \includegraphics[width=0.9\linewidth]{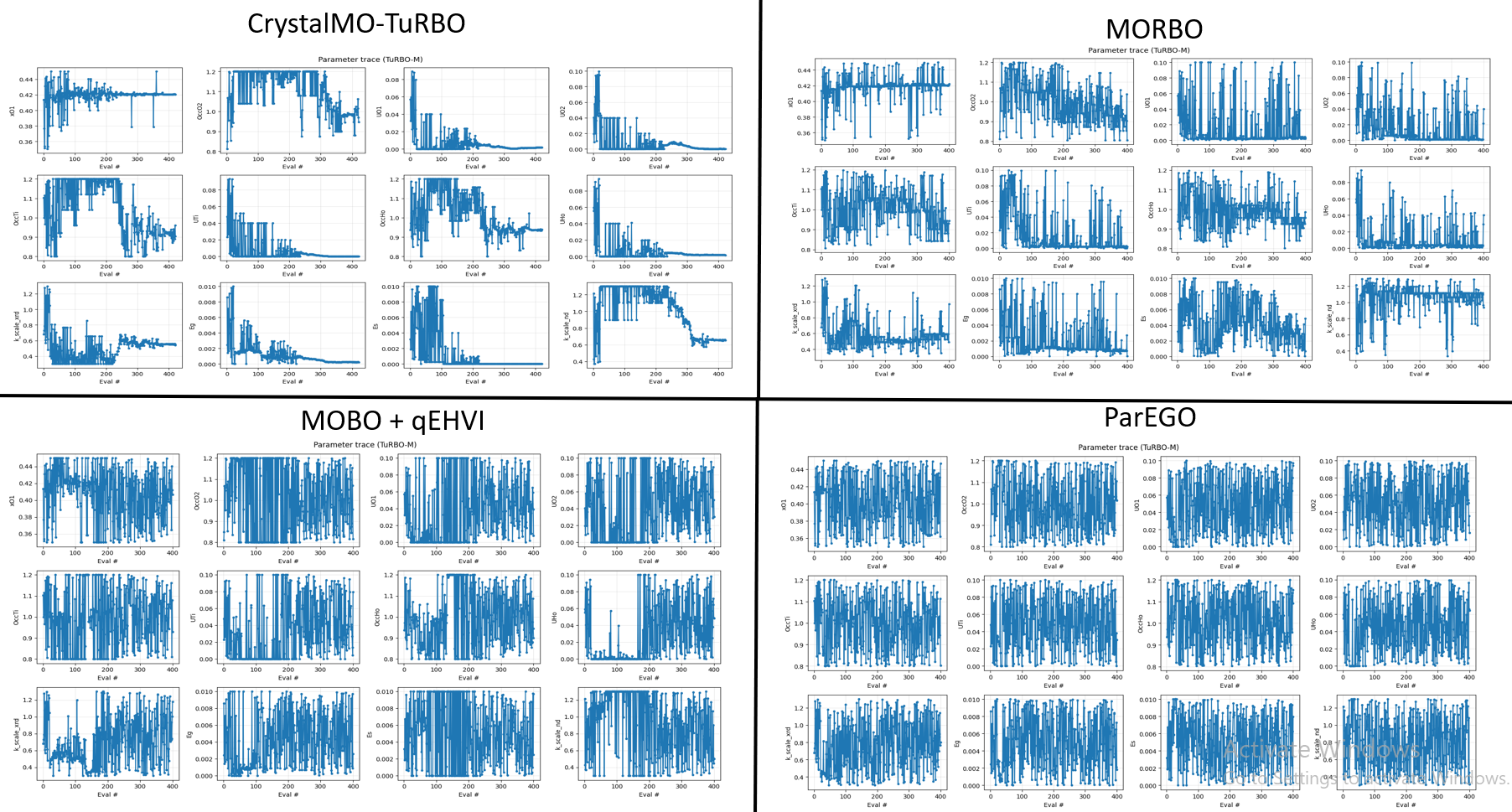}
    \caption{Figure showing the convergence of individual parameters for \emph{CrystalMO-TuRBO} and the other multi-objective BO baselines. As seen, in the phase 2 of the proposed \emph{CrystalMO-TuRBO} converged to some very small promising parameter space while other methods, especially qEHVI+MOBO and ParEGO continue to explore the full psrsmeter space. This correspond to the observation in Figure~\ref{fig:Fig11-gp_map}}
    \label{fig:Fig10-params_conv.png}
\end{figure}

\begin{figure}[t]
    \centering
    \includegraphics[width=\linewidth]{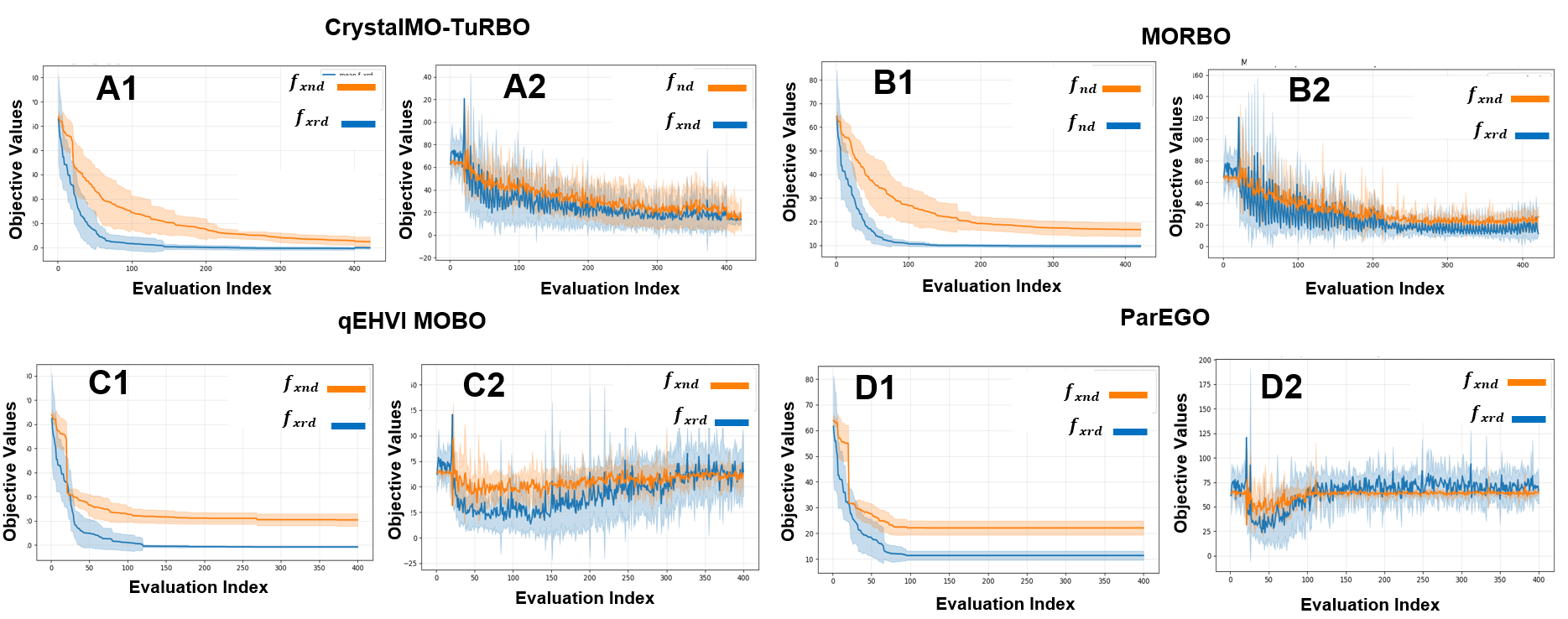}
    \caption{Figures showing convergence curve for the multiple run of the experiments. Values plotted are means and standard deviations of objective values. Column 1 and 2 for each methods are running minimumns and raw objective values respectively.}
    \label{fig:multi-run_comparison_all}
    \vspace{-2mm}
\end{figure}

Figures~\ref{fig:Fig13-pareto}, ~\ref{fig:Fig12-conv_curve}, ~\ref{fig:Fig11-gp_map} and ~\ref{fig:Fig10-params_conv.png} provide insight into the optimization behavior. The Pareto fronts (Figure~\ref{fig:Fig13-pareto} show that the TuRBO-based methods simultaneously reduce both objectives, whereas qEHVI and ParEGO primarily improve the XRD objective with comparatively limited improvement in the ND objective. Furthermore, the Pareto solutions obtained by CMT are more tightly clustered than those of MORBO, indicating greater robustness and refinement precision. The convergence curves (Figure~\ref{fig:Fig12-conv_curve}) illustrate the characteristic two-phase behavior of CMT, consisting of rapid global exploration followed by high-precision local refinement. The Gaussian process maps (Figure~\ref{fig:Fig11-gp_map}) show that CMT and MORBO successfully identify the most promising regions of the parameter space, whereas qEHVI and ParEGO exhibit poorer localization. Finally, the parameter convergence plots (Figure~\ref{fig:Fig10-params_conv.png}) show that most structural parameters converge during Phase~1, while Phase~2 primarily refines the occupancy parameters, demonstrating the complementary roles of the two optimization stages. To evaluate robustness, we repeated every experiment in Table~\ref{table:pe-se-combined} for average of 12 times using different random seeds. To isolate the contribution of the proposed Phase~2 refinement, both the Phase~1-only implementation and the complete CMT framework were allocated the same budget of 400 evaluations. The repeated-run results (Tables~\ref{tab:table1} and~\ref{tab:table2}) closely mirror the single-run results, confirming the robustness of the proposed method. In both primary and secondary extinction experiments, CMT with the 10\% local bounding box achieved the best overall performance, yielding combined objective values of 23.79 and 23.55, respectively. The repeated-run analysis also confirms the significant contribution of the proposed Phase~2 refinement strategy, with the largest improvements observed for the 5\% and 10\% local bounding boxes. Pairwise comparisons using the Mann--Whitney U test confirmed that the improvements achieved by CMT over the competing methods are statistically significant at the 0.05 significance level. The convergence statistics shown in Figure~\ref{fig:multi-run_comparison} further demonstrate the robustness of CMT.

Unlike the competing methods, which converge prematurely to suboptimal solutions, CMT continues to improve throughout the optimization budget, suggesting that additional evaluations could yield further gains. Figure~\ref{fig:multi-run_comparison} further isolates the contribution of the proposed Phase~2 design by illustrating the additional improvement achieved after the global exploration stage.

\begin{table}[ht]
\caption{Results of parameter refinement using least-squares joint refinement, CrystalMO-TurBO (CMT) with different analysis options, and other multi-objective Bayesian optimization approaches under primary (PE) and secondary (SE) extinction correction.}
\label{table:pe-se-combined}
\small
\setlength{\tabcolsep}{3pt}
\begin{adjustbox}{width=\textwidth}
\begin{tabular}{@{}llllllllllllllll@{}}
\toprule
\multicolumn{16}{@{}l}{\textit{Subtable 1: Results of Refinements with Primary Extinction Correction (PE)}} \\
\midrule

AO & XO1 & OccO2 & UO1 & UO2 & OccTi & UTi & OccHo & UHo & $Scale_{XRD}$ & Ep & Eg & Es & $Scale_{ND}$ & $f_{XRD}$ & $f_{ND}$ \\
\midrule

LS (PE)
  & \textit{\color{blue}0.42030}
  & \textit{\color{blue}1.00418}
  & \textit{\color{blue}0.0008593}
  & \textit{\color{blue}0.000042}
  & \textit{\color{blue}0.83726}
  & \textit{\color{blue}0.0000}
  & \textit{\color{blue}1.02523}
  & \textit{\color{blue}0.001905}
  & \textit{\color{blue}0.62677}
  & \textit{\color{blue}0.00078725}
  & --- & ---
  & \textit{\color{blue}0.61809}
  & \textit{\color{blue}28.50}
  & \textit{\color{blue}28.74} \\
CMT 1\% Ph 1 & 0.419095 & 1.166344 & 0.004439 & 0.000 & 1.200 & 0.000 & 1.146837 & 0.000 & 0.355875 & 0.000988 & --- & --- & 1.300 & 11.43 & 17.66 \\
CMT 1\% Ph 2 & 0.419476 & 1.178007 & 0.004395 & 0.000 & 1.200 & 0.000 & 1.158305 & 0.00 & 0.352316 & 0.000998 & --- & --- & 1.300 & 11.47 & 17.44 \\
CMT 2\% Ph 1 & 0.421186 & 1.199497 & 0.005409 & 0.000 & 1.000 & 0.000 & 1.047999 & 0.000 & 0.425788 & 0.000440 & --- & --- & 1.022328 & 10.11 & 17.12 \\
CMT 2\% Ph 2 & 0.421572 & 1.126275 & 0.002427 & 0.000941 & 0.932819 & 0.001574 & 1.2000 & 0.00 & 0.356125 & 0.000428 & --- & --- & 0.808468 & 9.24 & 14.57 \\
CMT 5\% Ph 1 & 0.42061 & 0.88000 & 0.006463 & 0.000 & 0.944039 & 0.000 & 0.920393 & 0.00 & 0.539907 & 0.000635 & --- & --- & 1.300000 & 10.73 & 18.26 \\
CMT 5\% Ph 2 & 0.42109 & 0.91024 & 0.005622 & 0.003594 & 0.800 & 0.001646 & 0.8000 & 0.001926 & 0.763544 & 0.000585 & --- & --- & 1.084155 & 11.499 & 13.21 \\
CMT 10\% Ph 1 & 0.421813 & 1.2000 & 0.000735 & 0.0050 & 0.800 & 0.000 & 1.100 & 0.000 & 0.426290 & 0.001249 & --- & --- & 1.30000 & 9.36 & 19.42 \\
\textit{\color{red}CMT 10\% Ph 2}
  & \textit{\color{red}0.420578}
  & \textit{\color{red}0.865217}
  & \textit{\color{red}0.004230}
  & \textit{\color{red}0.001588}
  & \textit{\color{red}0.800}
  & \textit{\color{red}0.000074}
  & \textit{\color{red}0.800}
  & \textit{\color{red}0.001299}
  & \textit{\color{red}0.759669}
  & \textit{\color{red}0.000393}
  & --- & ---
  & \textit{\color{red}0.91974}
  & \textit{\color{red}11.46}
  & \textit{\color{red}11.67} \\
\midrule
qEHVI MOBO & 0.42118 & 1.200 & 0.000 & 0.000 & 0.979665 & 0.000 & 1.200 & 0.000 & 0.3417596 & 0.00091355 & --- & --- & 1.3000 & 9.359 & 22.096 \\
ParEGO & 0.422222 & 0.814815 & 0.001174 & 0.004374 & 0.800 & 0.000 & 0.800 & 0.006882 & 0.869684 & 0.001192 & --- & --- & 1.3000 & 16.549 & 20.491 \\
MORBO & 0.420804 & 0.813548 & 0.004109 & 0.001127 & 0.819512 & 0.001145 & 0.843298 & 0.001429 & 0.693709 & 0.001053 & --- & --- & 1.272194 & 10.69 & 15.00 \\
\midrule
\multicolumn{16}{@{}l}{\textit{Subtable 2: Results of Refinements with Secondary Extinction Correction (SE)}} \\
\midrule
AO & XO1 & OccO2 & UO1 & UO2 & OccTi & UTi & OccHo & UHo & $Scale_{XRD}$ & Ep & Eg & Es & $Scale_{ND}$ & $f_{XRD}$ & $f_{ND}$ \\
\midrule
SP (SE)
  & \textit{\color{blue}0.42014}
  & \textit{\color{blue}0.974162}
  & \textit{\color{blue}0.005825}
  & \textit{\color{blue}0.003296}
  & \textit{\color{blue}0.912338}
  & \textit{\color{blue}0.002976}
  & \textit{\color{blue}0.950615}
  & \textit{\color{blue}0.003868}
  & \textit{\color{blue}0.62677}
  & ---
  & \textit{\color{blue}0.001277}
  & \textit{\color{blue}0.001651}
  & \textit{\color{blue}0.618089}
  & \textit{\color{blue}14.69}
  & \textit{\color{blue}41.14} \\
CMT 1\% Ph 1 & 0.421651 & 1.200 & 0.006327 & 0.00 & 0.800 & 0.00 & 1.2000 & 0.00 & 0.355657 & --- & 0.000839 & 0.008 & 1.189545 & 10.14 & 19.78 \\
CMT 1\% Ph 2 & 0.421651 & 1.200 & 0.006327 & 0.00 & 0.800 & 0.00 & 1.2000 & 0.00 & 0.355657 & --- & 0.000839 & 0.008 & 1.189545 & 10.14 & 19.78 \\
CMT 2\% Ph 1 & 0.419439 & 1.194077 & 0.00145 & 0.005622 & 0.870292 & 0.000364 & 1.152057 & 0.003446 & 0.391321 & --- & 0.001506 & 0.001039 & 1.278563 & 10.14 & 19.78 \\
CMT 2\% Ph 2 & 0.421132 & 0.975116 & 0.00514 & 0.005185 & 0.800 & 0.000409 & 0.8000 & 0.002088 & 0.764507 & --- & 0.000513 & 0.001753 & 1.036075 & 11.66 & 12.77 \\
CMT 5\% Ph 1 & 0.420975 & 1.1600 & 0.003147 & 0.000 & 0.950064 & 0.000 & 0.899091 & 0.002939 & 0.595342 & --- & 0.000463 & 0.000001 & 0.913139 & 12.36 & 14.57 \\
CMT 5\% Ph 2 & 0.420779 & 1.1400 & 0.000565 & 0.004423 & 1.083000 & 0.004265 & 0.989469 & 0.001457 & 0.486170 & --- & 0.000201 & 0.000001 & 0.581740 & 11.51 & 10.65 \\
CMT 10\% Ph 1 & 0.419048 & 1.200 & 0.005297 & 0.000 & 1.2000 & 0.000 & 1.117742 & 0.00 & 0.358987 & --- & 0.000836 & 0.000001 & 1.2000 & 11.38 & 17.66 \\
\textit{\color{red}CMT 10\% Ph 2}
  & \textit{\color{red}0.421250}
  & \textit{\color{red}0.944536}
  & \textit{\color{red}0.001328}
  & \textit{\color{red}0.000445}
  & \textit{\color{red}0.913575}
  & \textit{\color{red}0.000134}
  & \textit{\color{red}0.948550}
  & \textit{\color{red}0.001979}
  & \textit{\color{red}0.549276}
  & ---
  & \textit{\color{red}0.000247}
  & \textit{\color{red}0.000001}
  & \textit{\color{red}0.65694}
  & \textit{\color{red}10.36}
  & \textit{\color{red}10.46} \\
\midrule
qEHVI MOBO & 0.421443 & 1.200 & 0.005029 & 0.000 & 0.884216 & 0.002369 & 1.173446 & 0.000 & 0.372995 & --- & 0.000672 & 0.000001 & 1.070952 & 9.33 & 16.56 \\
ParEGO & 0.41578 & 0.800 & 0.000 & 0.095920 & 0.8000 & 0.000 & 1.192112 & 0.000 & 0.404053 & --- & 0.000745 & 0.007124 & 1.300 & 13.688 & 32.949 \\
MORBO & 0.42021 & 0.903372 & 0.004712 & 0.001434 & 0.843886 & 0.001312 & 0.925901 & 0.002807 & 0.591638 & --- & 0.000728 & 0.000030 & 1.11580 & 10.968 & 14.057 \\
\bottomrule
\end{tabular}
\end{adjustbox}
\end{table}

\begin{table*}[t]
\centering

\begin{minipage}[t]{0.49\textwidth}
\centering
\caption{Repeated Run Results for Multi-objective BO in Sub-table 1 of Table~\ref{table:pe-se-combined}.}
\label{tab:table1}
\small
\setlength{\tabcolsep}{3pt}
\resizebox{\linewidth}{!}{
\begin{tabular}{@{}lcccc@{}}
\toprule
Analysis Options & $N_{\mathrm{Iterations}}$ & $f_{xrd}$ & $f_{nd}$ & $f_{sum}$ \\
\midrule
LS & 400 & $42.47 \pm 24.15$ & $59.89 \pm 4.67$ & $102.35 \pm 27.48$ \\
\midrule
CMT 1\% Ph1 & 400 & $10.82 \pm 1.06$ & $17.87 \pm 2.51$ & $28.69 \pm 2.43$ \\
CMT 1\% Ph1+Ph2 & 400 & $12.09 \pm 1.50$ & $15.98 \pm 2.43$ & $28.07 \pm 2.53$ \\
CMT 2\% Ph1 & 400 & $10.82 \pm 1.06$ & $17.87 \pm 2.51$ & $28.69 \pm 2.43$ \\
CMT 2\% Ph1+Ph2 & 400 & $12.15 \pm 1.93$ & $15.10 \pm 2.21$ & $27.26 \pm 2.86$ \\
CMT 5\% Ph1 & 400 & $10.81 \pm 1.06$ & $17.87 \pm 2.51$ & $28.69 \pm 2.43$ \\
CMT 5\% Ph1+Ph2 & 400 & $11.07 \pm 1.09$ & $14.66 \pm 3.08$ & $25.72 \pm 3.06$ \\
CMT 10\% Ph1 & 400 & $10.81 \pm 1.06$ & $17.87 \pm 2.50$ & $28.69 \pm 2.47$ \\
\textit{\color{red}CMT 10\% Ph1+Ph2} & \textit{\color{red}400} & \textit{\color{red}$10.48 \pm 1.06$} & \textit{\color{red}$13.31 \pm 2.19$} & \textit{\color{red}$23.79 \pm 2.11$} \\
\midrule
qEHVI + MOBO & 400 & $10.86 \pm 1.02$ & $21.88 \pm 3.01$ & $32.74 \pm 3.16$ \\
ParEGO & 400 & $12.52 \pm 1.75$ & $22.48 \pm 2.82$ & $35.04 \pm 4.08$ \\
MORBO & 400 & $10.70 \pm 1.06$ & $18.06 \pm 2.44$ & $28.77 \pm 2.51$ \\
\bottomrule
\end{tabular}
}
\end{minipage}
\hfill
\begin{minipage}[t]{0.49\textwidth}
\centering
\caption{Repeated Run Results for Multi-objective BO in Sub-table 2 of Table~\ref{table:pe-se-combined}}
\label{tab:table2}
\small
\setlength{\tabcolsep}{3pt}
\resizebox{\linewidth}{!}{
\begin{tabular}{@{}lcccc@{}}
\toprule
Analysis Options & $N_{\mathrm{Iterations}}$ & $f_{xrd}$ & $f_{nd}$ & $f_{sum}$ \\
\midrule
LS & 400 & $44.57 \pm 23.63$ & $52.68 \pm 11.03$ & $97.24 \pm 32.19$ \\
\midrule
CMT 1\% Ph1 & 400 & $11.19 \pm 1.77$ & $16.62 \pm 3.12$ & $27.81 \pm 2.50$ \\
CMT 1\% Ph1+Ph2 & 400 & $10.27 \pm 1.06$ & $17.68 \pm 1.60$ & $27.95 \pm 1.75$ \\
CMT 2\% Ph1 & 400 & $11.19 \pm 1.78$ & $16.47 \pm 3.24$ & $27.66 \pm 2.65$ \\
CMT 3\% Ph1+Ph2 & 400 & $10.23 \pm 0.66$ & $17.29 \pm 1.98$ & $27.52 \pm 2.08$ \\
CMT 5\% Ph1 & 400 & $11.18 \pm 1.78$ & $16.47 \pm 3.24$ & $27.65 \pm 2.66$ \\
CMT 5\% Ph1+Ph2 & 400 & $10.27 \pm 0.60$ & $16.51 \pm 3.05$ & $26.78 \pm 2.83$ \\
CMT 10\% Ph1 & 400 & $11.19 \pm 1.78$ & $16.47 \pm 3.24$ & $27.65 \pm 2.68$ \\
\textit{\color{red}CMT 10\% Ph1+Ph2} & \textit{\color{red}400} & \textit{\color{red}$11.09 \pm 1.46$} & \textit{\color{red}$12.46 \pm 2.51$} & \textit{\color{red}$23.55 \pm 3.03$} \\
\midrule
qEHVI + MOBO & 400 & $10.17 \pm 6.89$ & $18.37 \pm 1.82$ & $28.85 \pm 1.57$ \\
ParEGO & 400 & $13.63 \pm 2.29$ & $22.16 \pm 2.75$ & $35.27 \pm 3.01$ \\
MORBO & 400 & $11.49 \pm 1.83$ & $16.44 \pm 3.49$ & $27.93 \pm 2.73$ \\
\bottomrule
\end{tabular}
}
\end{minipage}

\end{table*}

\subsection{Empirical Justification for \textit{CrystalMO-TuRBO}}

The motivation for formulating crystal structure refinement as a multi-objective optimization problem is illustrated by the single-objective refinement results in Tables~\ref{table4} and~\ref{table5}. Although both XRD and ND characterize the same crystal, optimizing each modality independently produces substantially different optimal parameter sets. For example, XRD refinement estimates the holmium occupancy as $Occ{\mathrm{Ho}}=1.100$, suggesting excess holmium, whereas ND refinement estimates $Occ{\mathrm{Ho}}=0.827$, implying significant holmium deficiency. Similar inconsistencies are observed for the Ti and O occupancies as well as several atomic displacement parameters. These contradictory solutions arise because XRD and ND probe complementary physical properties and therefore provide different optimization landscapes. The conventional least-squares joint refinement resolves this conflict by simultaneously optimizing the two objectives, producing a single compromise solution. However, this compromise is obtained at the expense of substantially larger residual errors. For the primary extinction experiment, least-squares refinement achieves $f_{\mathrm{XRD}}=28.50$ and $f_{\mathrm{ND}}=28.74$, considerably worse than the independently optimized single-objective solutions ($8.82$ and $11.02$, respectively). These observations are consistent with the limitations of conventional scalarized refinement discussed in Section~\ref{sec:related_work} and motivate a true multi-objective formulation. The proposed \emph{CMT} addresses this limitation by jointly optimizing both diffraction objectives by searching for the best compromise between the 2 objectives. It identifies a physically consistent parameter set while achieving objective values of $f_{\mathrm{XRD}}=11.46$ and $f_{\mathrm{ND}}=11.67$, approaching the performance of the corresponding single-objective refinements and substantially outperforming least-squares joint refinement. Similar trends are consistently observed across the remaining experiments. Beyond identifying a high-quality compromise solution, \emph{CMT} also constructs an empirical Pareto frontier and learns Gaussian process surrogate models of the refinement landscape. These additional outputs provide insight into the trade-offs between XRD and ND refinement and characterize the underlying parameter space, capabilities that are not available in conventional least-squares refinement. This richer representation of the refinement problem enables both improved optimization performance and a deeper understanding of the competing experimental objectives.

\subsection{Ablation Study}

We performed ablation studies to evaluate the effects of the proposed local bounding-box hyperparameter and the Phase~2 local refinement strategy. Tables~\ref{tab:table1} and~\ref{tab:table2} show that the performance of \emph{CrystalMO-TuRBO} is sensitive to the bounding-box width. Specifically, performance improves as the box width increases from 1\% to 10\%, with the 10\% bounding box consistently achieving the best overall refinement accuracy. Smaller bounding boxes are overly restrictive and may exclude nearby high-quality solutions during Phase~2, whereas excessively large boxes diminish the benefit of localized high-precision refinement. These results indicate that careful selection of the bounding-box width is important for achieving optimal performance.

To isolate the contribution of the proposed Phase~2 design, we compared CMT using only Phase~1 with the complete Phase~1 + Phase~2 framework, while allocating the same budget of 400 function evaluations to both implementations. As shown in Tables~\ref{tab:table1} and~\ref{tab:table2}, incorporating Phase~2 consistently improves refinement performance, with the largest gains observed for the 5\% and 10\% bounding-box settings, where the best overall solutions were obtained. This improvement is further illustrated by the convergence curves in Figure~\ref{fig:multi-run_comparison}, which show that Phase~2 continues to reduce the objective values after the global exploration stage has converged.


\subsection{Limitations}
\label{limitation}

A limitation of this study is that the empirical evaluation is conducted on X-ray and neutron diffraction data obtained from a single high-quality $Ho_{2}Ti_{2}O_{7}$ crystal. This choice was deliberate because acquiring matched single-crystal X-ray and neutron diffraction datasets from the same specimen is experimentally expensive and time-consuming. Using paired datasets from a single crystal eliminates inter-sample variability and ensures that both optimization objectives correspond to the same underlying crystal structure, providing a controlled setting for evaluating joint refinement methods. Although the evaluation is limited to a single material system, the optimization problem remains highly challenging due to the intrinsic characteristics of crystal structure refinement, including expensive function evaluations, strong parameter correlations, non-convex optimization landscapes, and conflicting information from complementary diffraction modalities. These challenges arise from the refinement problem itself rather than from a particular crystal sample. Consequently, the proposed framework is expected to be applicable to other crystallographic refinement problems involving multiple complementary diffraction datasets. However, the empirical claims of this work are limited to the $Ho_{2}Ti_{2}O_{7}$ system evaluated in this study.
An important direction for future work is to evaluate the proposed framework on additional crystals, material systems, and other scientific inverse problems requiring joint optimization of multiple complementary experimental objectives. Such studies would further establish the generality and robustness of the proposed approach under different materials, experimental conditions, and defect configurations.
\section{Conclusion}

We presented \textit{CrystalMO-TuRBO} (CMT), a multi-objective trust-region Bayesian optimization framework for joint crystal structure refinement using complementary X-ray and neutron diffraction data. Unlike conventional least-squares joint refinement, which optimizes a scalarized objective, the proposed framework formulates refinement as a true multi-objective optimization problem, enabling simultaneous optimization of both diffraction modalities while preserving their complementary information. The proposed two-phase trust-region design combines global exploration with high-precision local refinement, making it particularly well suited for scientific inverse problems requiring extremely accurate parameter estimation. Extensive experiments on the $Ho_{2}Ti_{2}O_{7}$ system demonstrate that CMT consistently outperforms the crystallographic community standard least-squares joint refinement workflow as well as representative multi-objective Bayesian optimization baselines, including qEHVI, ParEGO, and MORBO. Repeated-run experiments further confirm the robustness of the proposed framework, the statistically significant contribution of the Phase~2 local refinement strategy, and the importance of the local bounding-box hyperparameter in achieving high-precision refinement. Beyond improved optimization performance, the proposed framework provides richer scientific insight by constructing an empirical Pareto frontier and learning surrogate models of the refinement landscape, capabilities that are not available in conventional least-squares refinement. Although evaluated on a single $Ho_{2}Ti_{2}O_{7}$ crystal in this work, the proposed formulation is broadly applicable to expensive multi-modal inverse problems where multiple complementary data sources must be jointly reconciled. We hope this work encourages broader adoption of multi-objective Bayesian optimization for scientific discovery and materials characterization.

\clearpage


\bibliographystyle{plainnat}
\bibliography{references}


\newpage

\appendix

\clearpage
\section{Tables}

\begin{table}[h]
\caption{Results of parameter refinement using the least-square based joint refinement, the proposed CrystalMO-TurBO (CMT) with different analysis options, and other multi-objective Bayesian optimization approaches with primary extinction correction}
\label{table1}
\small
\setlength{\tabcolsep}{4pt}
\begin{adjustbox}{width=\textwidth}
\begin{tabular}{@{}llllllllllllll@{}}

\toprule

AO & XO1 & OccO2 & UO1 & UO2 & OccTi & UTi & OccHo & UHo &  $Scale_{XRD}$ & Ep & $Scale_{ND}$ & $f_{XRD}$ & $f_{ND}$ \\

\midrule

LS (PE) & \textit{\color{brown}0.42030} & \textit{\color{brown}1.00418} & \textit{\color{brown}0.0008593} & \textit{\color{brown}0.000042} & \textit{\color{brown}0.83726} & \textit{\color{brown}0.0000} & \textit{\color{brown}1.02523} & \textit{\color{brown}0.001905} & \textit{\color{brown}0.62677}& \textit{\color{brown}0.00078725} & \textit{\color{brown}0.61809} & \textit{\color{brown}28.50} & \textit{\color{brown}28.74} \\

CMT 1\% Ph 1 & 0.419095 & 1.166344 & 0.004439 & 0.000 & 1.200 & 0.000 & 1.146837 & 0.000 & 0.355875 & 0.000988 & 1.300 &  11.43 & 17.66 \\

CMT 1\% Ph 2 & 0.419476 & 1.178007 & 0.004395 & 0.000 & 1.200 & 0.000 & 1.158305 & 
0.00 & 0.352316 & 0.000998 & 1.300  & 11.47  & 17.44 \\

CMT 2\% Ph 1 & 0.421186 & 1.199497 & 0.005409 & 0.000 & 1.000 & 0.000 & 1.047999 & 0.000 & 0.425788 & 0.000440 & 1.022328 &  10.11 & 17.12 \\

CMT 2\% Ph 2 & 0.421572 & 1.126275 & 0.002427 & 0.000941 & 0.932819 & 0.001574 & 1.2000 & 
0.00 & 0.356125 & 0.000428 & 0.808468  & 9.24  & 14.57 \\

CMT 5\% Ph 1 & 0.42061 & 0.88000 & 0.006463 & 0.000 & 0.944039 & 0.000 & 0.920393 & 
0.00 & 0.539907 & 0.000635 & 1.300000  & 10.73  & 18.26\\

CMT 5\% Ph 2 & 0.42109 & 0.91024 & 0.005622 & 0.003594 & 0.800 & 0.001646 & 0.8000 & 0.001926 & 0.763544 & 0.000585 & 1.084155 &  11.499 & 13.21 \\

CMT 10\% Ph 1 & 0.421813 & 1.2000 & 0.000735 & 0.0050 & 0.800 & 0.000 & 1.100 & 0.000 & 0.426290 & 0.001249 & 1.30000 &  9.36 & 19.42 \\

\textit{\color{red}CMT 10\% Ph 2} & \textit{\color{red}0.420578} & \textit{\color{red}0.865217} & \textit{\color{red}0.004230} & \textit{\color{red}0.001588} & \textit{\color{red}0.800} & \textit{\color{red}0.000074} & \textit{\color{red}0.800} & 
\textit{\color{red}0.001299} & \textit{\color{red}0.759669} & \textit{\color{red}0.000393} & \textit{\color{red}0.91974}  & \textit{\color{red}11.46}  & \textit{\color{red}11.67} \\

\midrule

qEHVI MOBO & 0.42118 & 1.200 & 0.000 & 0.000 & 0.979665 & 0.000 & 1.200 & 0.000 & 0.3417596 & 0.00091355 & 1.3000 &  9.359 & 22.096 \\

ParEGO & 0.422222 & 0.814815 & 0.001174 & 0.004374 & 0.800 & 0.000 & 0.800& 0.006882 & 0.869684 & 0.001192 & 1.3000 &  16.548874 & 20.490881 \\

MORBO & 0.420804 & 0.813548 & 0.004109 & 0.001127 & 0.819512 & 0.001145 & 0.843298 & 0.001429 & 0.693709 & 0.001053 & 1.272194 &  10.69 & 15.00 \\

\bottomrule
\end{tabular}
\end{adjustbox}
\end{table}

\begin{table}[h]
\caption{Results of parameter refinement using the least-square based joint refinement, the proposed CrystalMO-TurBO (CMT) with different analysis options, and other multi-objective Bayesian optimization approaches with secondary extinction correction}
\label{table2}
\small
\small
\setlength{\tabcolsep}{4pt}
\begin{adjustbox}{width=\textwidth}
\begin{tabular}{@{}llllllllllllllll@{}}

\toprule

AO & XO1 & OccO2 & UO1 & UO2 & OccTi & UTi & OccHo & UHo &  $Scale_{XRD}$ & $Eg$ & $Es$ & $Scale_{ND}$ & $f_{XRD}$ & $f_{ND}$ \\

\midrule
SP (SE) & \textit{\color{brown}0.42014} & \textit{\color{brown}0.974162} & \textit{\color{brown}0.005825} & \textit{\color{brown}0.003296} & \textit{\color{brown}0.912338} & \textit{\color{brown}0.002976} & \textit{\color{brown}0.950615} & \textit{\color{brown}0.003868} & \textit{\color{brown}0.62677}& \textit{\color{brown}0.001277} & 
\textit{\color{brown}0.001651} &
\textit{\color{brown}0.618089} & \textit{\color{brown}14.69} & \textit{\color{brown}41.14} \\


CMT 1\% Ph 1 & 0.421651 & 1.200& 0.006327 & 0.00 & 0.800 & 0.00 & 1.2000 & 0.00 & 0.355657 & 0.000839 & 0.008 &  1.189545 & 10.14 & 19.78 \\

CMT 1\% Ph 2 & 0.421651 & 1.200& 0.006327 & 0.00 & 0.800 & 0.00 & 1.2000 & 0.00 & 0.355657 & 0.000839 & 0.008 &  1.189545 & 10.14 & 19.78 \\

CMT 2\% Ph 1 & 0.419439 & 1.194077 & 0.00145 & 0.005622 & 0.870292 & 0.000364 & 1.152057 & 0.003446 & 0.391321 & 0.001506 & 0.001039 &  1.278563  & 10.14 & 19.78 \\

CMT 2\% Ph 2 & 0.421132 & 0.975116 & 0.00514 & 0.005185 & 0.800 & 0.000409 & 0.8000 & 0.002088 & 0.764507 & 0.000513 & 0.001753 &  1.036075 & 11.66 & 12.77 \\

CMT 5\% Ph 1 & 0.420975 & 1.1600 & 0.003147 & 0.000 & 0.950064 & 0.000 & 0.899091 & 0.002939 & 0.595342 & 0.000463 & 0.000001 &  0.913139  & 12.36 & 14.57 \\

CMT 5\% Ph 2 & 0.420779 & 1.1400 & 0.000565 & 0.004423 & 1.083000 & 0.004265 & 0.989469 & 0.001457 & 0.486170 & 0.000201 & 0.000001 &  0.581740 & 11.51 & 10.65 \\

CMT 10\% Ph 1 & 0.419048 & 1.200 & 0.005297 & 0.000 & 1.2000 & 0.000 & 1.117742 & 0.00 & 0.358987 & 0.000836 & 0.000001 &  1.2000  & 11.38 & 17.66 \\

\textit{\color{red}CMT 10\% Ph 2} & \textit{\color{red}0.421250} & \textit{\color{red}0.944536} & \textit{\color{red}0.001328} & \textit{\color{red}0.000445} & \textit{\color{red}0.913575} & \textit{\color{red}0.000134} & \textit{\color{red}0.948550} & \textit{\color{red}0.001979} & \textit{\color{red}0.549276} & \textit{\color{red}0.000247} & \textit{\color{red}0.000001} &  \textit{\color{red}0.65694} & \textit{\color{red}10.36} & \textit{\color{red}10.46} \\

\midrule

qEHVI MOBO & 0.421443 & 1.200 & 0.005029 & 0.000 & 0.884216 & 0.002369 & 1.173446  & 0.000 & 0.372995 & 0.000672 & 0.000001 & 1.070952 & 9.33 & 16.56 \\

ParEGO & 0.41578 & 0.800 & 0.000 & 0.095920 & 0.8000 & 0.000 & 1.1921122  & 0.000 & 0.404053 & 0.00074488 & 0.007124168 & 1.300 & 13.688 & 32.949 \\

MORBO & 0.42021 & 0.903372 & 0.004712 & 0.001434 & 00.843886 & 0.001312 & 0.9259014  & 0.0028073 & 0.591638 & 0.00072765 & 0.0000304 & 1.11580 & 10.968 & 14.057 \\

\bottomrule
\end{tabular}
\end{adjustbox}
\end{table}

\begin{table}[h]
\caption{Here, we implemented single objective Bayesian optimization for refinement of structure parameters of the same crystal using X-ray diffraction data. We impleneted the BO workflow proposed by \cite{Agada2025BORefinement}. but with scale refined as BO parameter instead of having scale refined in GSAS-II}
\label{table4}
\small
\setlength{\tabcolsep}{4pt}
\begin{adjustbox}{width=\textwidth}
\begin{tabular}{@{}llllllllllll@{}}

\toprule

Analysis Options & XO1 & OccO2 & UO1 & UO2 & OccTi & UTi & OccHo & UHo & $k_scale$ & $f_{XRD}$ \\

\midrule
LS Refinement & \textit{\color{brown}0.42161} & \textit{\color{brown}0.92172} & \textit{\color{brown}0.008275} & \textit{\color{brown}0.00318} & \textit{\color{brown}0.80} & \textit{\color{brown}0.00} & \textit{\color{brown}0.9169} & \textit{\color{brown}0.003293} & \textit{\color{brown}0.660} & \textit{\color{brown}37.63} \\

\midrule

6-D BO (XRD) & 0.421597 & 0.920386 & 0.008262 & 0.00325 & 1.014632 & 0.001014 & 1.044677 & 0.000710 & 0.442897 & 9.9196 \\

10-D BO (XRD) & 0.360180 & 1.058103 & 0.074479 & 0.056498 & 0.921082 & 0.005700 & 1.002813 & 0.001931 & 0.496593 & 14.4501 \\

SAASBO (XRD) & 0.424384 & 0.937295 & 0.011161 & 0.003388 & 0.944646 & 0.002113 & 1.024816 & 0.000778 & 0.477134 & 9.4760 \\

\textit{\color{red}turbo-1 (XRD)} & \textit{\color{red}0.425274} & \textit{\color{red}1.014146} & \textit{\color{red}0.000000} & \textit{\color{red}0.000000} & \textit{\color{red}0.900000} & \textit{\color{red}0.000000} & \textit{\color{red}1.100000} & \textit{\color{red}0.001294} & \textit{\color{red}0.427654} & \textit{\color{red}8.8244} \\

\bottomrule
\end{tabular}
\end{adjustbox}
\end{table}

\begin{table}[h]
\caption{Here, we implemented single objective Bayesian optimization for refinement of structure parameters of the same crystal using neutron diffraction data, with primary extinction correction. The structure, scale and extinction parameters are all refined as BO parameters.}
\label{table5}
\small
\setlength{\tabcolsep}{4pt}
\begin{adjustbox}{width=\textwidth}
\begin{tabular}{@{}llllllllllll@{}}

\toprule

Analysis Options & XO1 & OccO2 & UO1 & UO2 & OccTi & UTi & OccHo & UHo & Ep & $k_scale$ & $f_{ND}$ \\

\midrule

LS Refinement (PE) & \textit{\color{brown}0.420438} & \textit{\color{brown}1.0326} & \textit{\color{brown}0.00} & \textit{\color{brown}0.00} & \textit{\color{brown}0.993181} & \textit{\color{brown}0.00} & \textit{\color{brown}0.961766} & \textit{\color{brown}0.00} & \textit{\color{brown}0.0001383} & \textit{\color{brown}0.399749} & \textit{\color{brown}42.21} \\

\midrule

7-D BO (ND PE) & 0.4200 & 1.005292 & 0.001002 & 0.005490 & 0.947391 & 0.000 & 0.972388 & 0.00 & 0.000058 & 0.398751 & 14.9109 \\

11-D BO (ND PE) & 0.425015 & 0.876277 & 0.005833 & 0.011765 & 1.181555 & 0.015559 & 1.134794 & 0.003597 & 0.000505 & 0.517131 & 36.89 \\

SAASBO (ND PE) & 0.422310 & 1.102841 & 0.002819 & 0.028820 & 0.853045 & 0.005746 & 0.928452 & 0.007735 & 0.000158 &  0.582048 & 25.29 \\

turbo-1 (ND PE) & \textit{\color{red}0.420408} & \textit{\color{red}1.198079} & \textit{\color{red}0.001151} & \textit{\color{red}0.002312} & \textit{\color{red}1.115125} & \textit{\color{red}0.003360} & \textit{\color{red}0.827218} & \textit{\color{red}0.000039} & \textit{\color{red}0.000181} & \textit{\color{red}0.599348} & \textit{\color{red}11.02} \\

\bottomrule
\end{tabular}
\end{adjustbox}
\end{table}

\begin{table}[h]
\caption{Here, we implemented single objective Bayesian optimization for refinement of structure parameters of the same crystal using neutron diffraction data, with secondary extinction correction. The structure, scale and extinction parameters are all refined as BO parameters}
\label{table6}
\small
\setlength{\tabcolsep}{4pt}
\begin{adjustbox}{width=\textwidth}
\begin{tabular}{@{}lllllllllllll@{}}

\toprule

Analysis Options & XO1 & OccO2 & UO1 & UO2 & OccTi & UTi & OccHo & UHo & Es & Eg & $k_scale$ & $f_{ND}$ \\

\midrule
LS Refinement (SE) & \textit{\color{brown}0.420457} & \textit{\color{brown}1.002304} & \textit{\color{brown}0.001} & \textit{\color{brown}0.001} & \textit{\color{brown}0.997867} & \textit{\color{brown}0.001} & \textit{\color{brown}0.971403} & \textit{\color{brown}0.001} & \textit{\color{brown}0.0017573} & \textit{\color{brown}0.000} & \textit{\color{brown}0.42629} & \textit{\color{brown}40.73} \\

\midrule

8-D BO (ND SE) & 0.420 & 1.1604 & 0.003341 & 0.008044 & 0.947391 & 0.0 & 0.972388 & 0.0 & 0.008797 & 0.002250 & 0.519675 & 11.8457 \\

12-D BO (ND SE) & 0.414975 & 1.004983 & 0.002483 & 0.003206 & 0.941292 & 0.006855 & 0.903200 & 0.010035 & 0.001513 & 0.003107 & 0.435911 & 30.16 \\

SAASBO (ND SE) & 0.423649 & 0.854168 & 0.000972 & 0.005691 & 1.092722 & 0.008958 & 1.087045 & 0.007280 & 0.008039 & 0.002349 & 0.476151 & 19.03 \\

turbo-1 (ND SE) & \textit{\color{red}0.420530} & \textit{\color{red}1.175704} & \textit{\color{red}0.002107} & \textit{\color{red}0.001502} & \textit{\color{red}1.174778} & \textit{\color{red}0.003085} & \textit{\color{red}1.189298} & \textit{\color{red}0.007244} & \textit{\color{red}0.009546} & \textit{\color{red}0.003286} & \textit{\color{red}0.583726} & \textit{\color{red}10.0388} \\

\bottomrule
\end{tabular}
\end{adjustbox}
\end{table}


\section{Figures for Multi-objective BO}

\begin{figure}[H]
    \centering
    \includegraphics[width=0.9\linewidth]{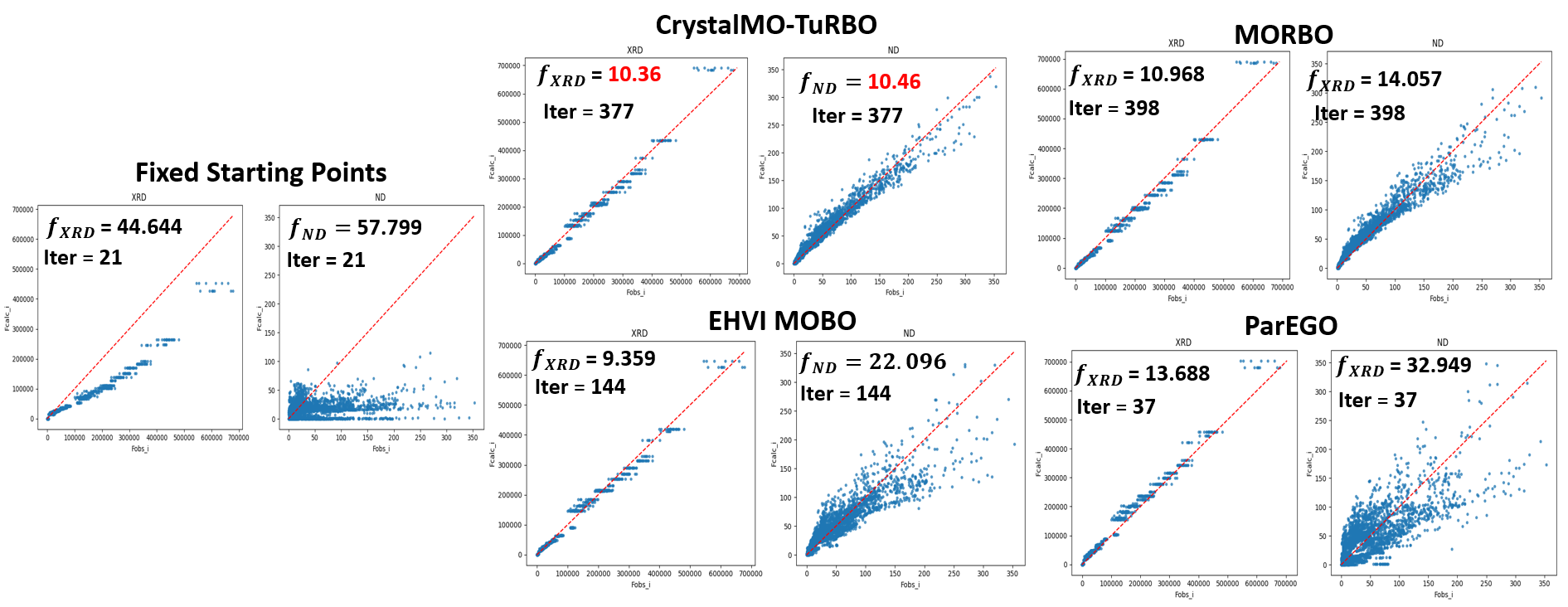}
    \caption{Figure showing scatter plots of the refined theoretical model $I_{cal}$ and the observed experimental observation $I_{obs}$ across different refinement methods and scattering data (X-ray and Neutron diffraction with secondary extinction correction). This plot visually depicts the mismatch of the experimental and theoretical reflections for the \emph{CrystalMO-TuRBO} and various multi-objective BO baselines such as MORBO \cite{Daulton2022MORBO}, ParEGO \cite{Knowles2006ParEGO} and \cite{Daulton2020qEHVI}. Reflections for one of the starting sets of parameters is added to portray how well each of the BO methods did against a completely non-optimal solution. As seen in the plot, \emph{CrystalMO-TurRBO} is the best performing, with high-dimensional MOBO as second.}
    \label{fig:fcal_vs_fobs_mobo}
\end{figure}

\section{Figures for Single objective BO}

\begin{figure}[H]
    \centering
    \includegraphics[width=0.9\linewidth]{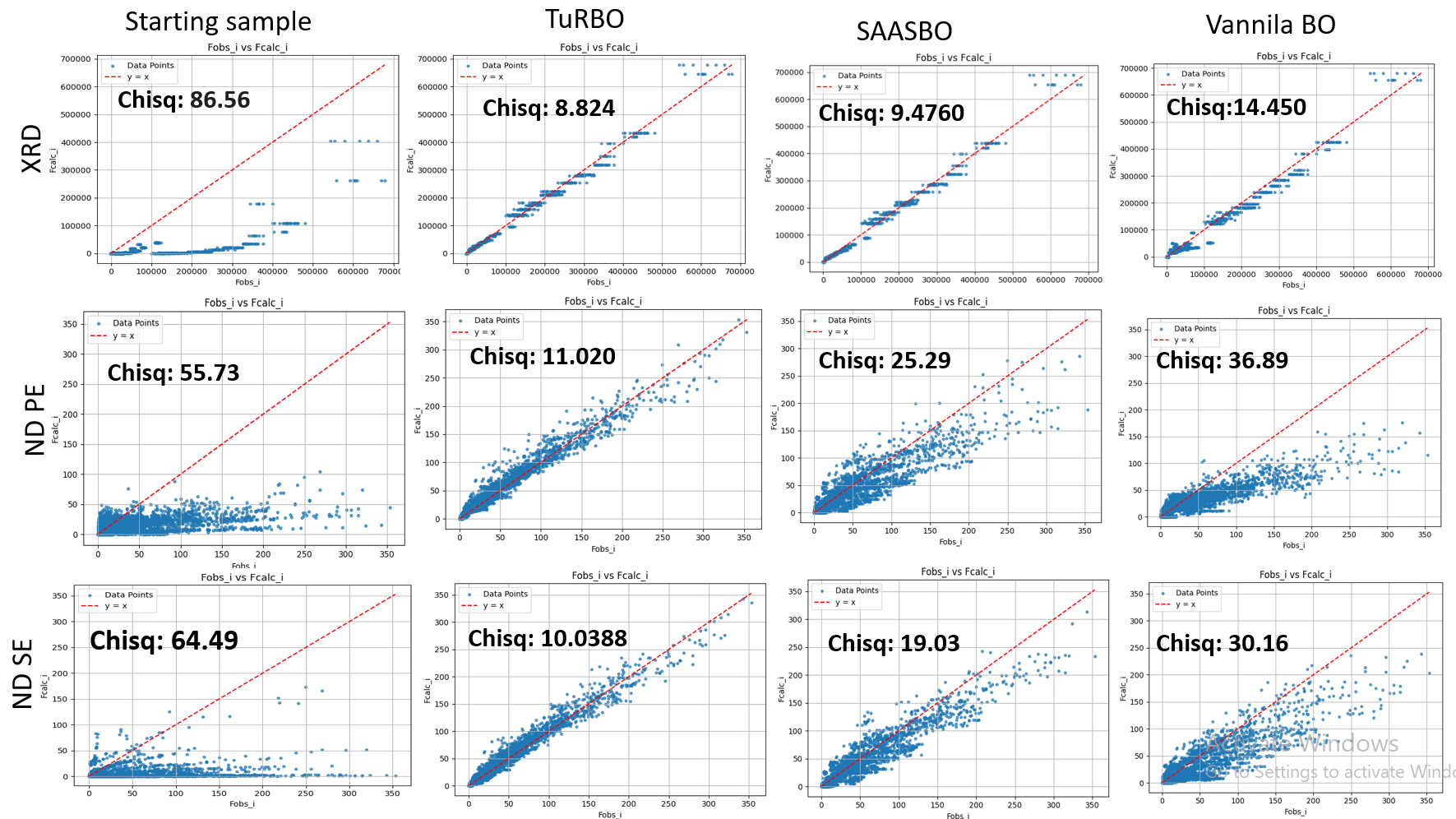}
    \caption{Figure showing scatter plots of the refined theoretical model $I_{cal}$ and the observed experimental observation $I_{obs}$ across one starting sample and different refinement methods (columns), and scattering data (rows). XRD mean refinement with X-ray diffraction data; ND PE means refinement using neutron diffraction with primary extinction correction; and ND SE means refinement using neutron diffraction data with secondary extinction.}
    \label{fig:Fig3-fcal_vs_fobs}
\end{figure}

\begin{figure}[H]
    \centering
    \includegraphics[width=0.9\linewidth]{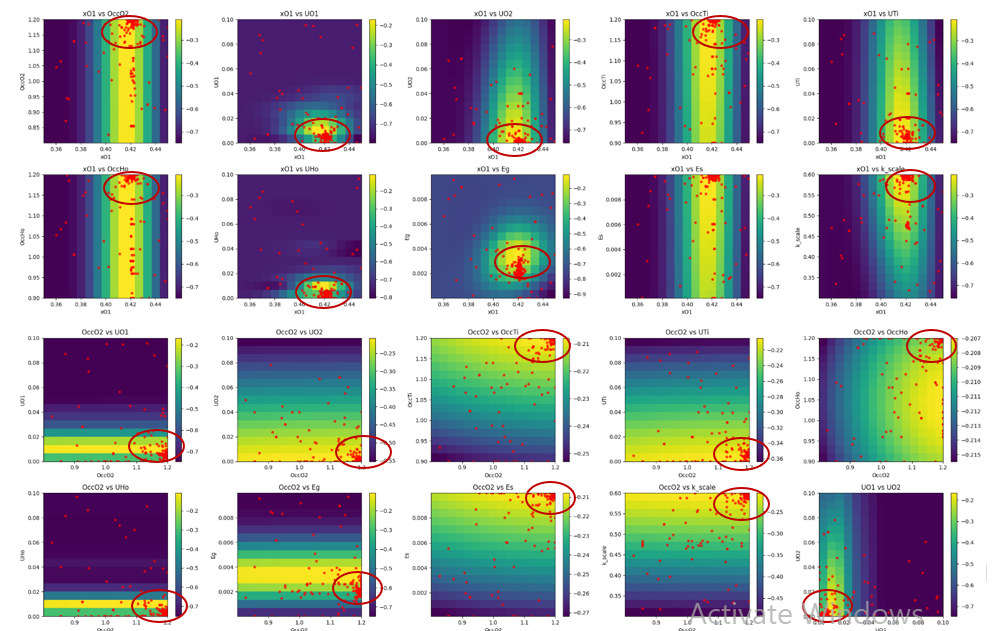}
    \caption{Figure showing the GP map means for the best performing TurBO implementation. As can be seen, the BO focuses more on the region of the parameter space where low $\chi^2$ is predicted. }
    \label{fig:Fig4-GP_Map_Mean}
\end{figure}

\begin{figure}[H]
    \centering
    \includegraphics[width=0.9\linewidth]{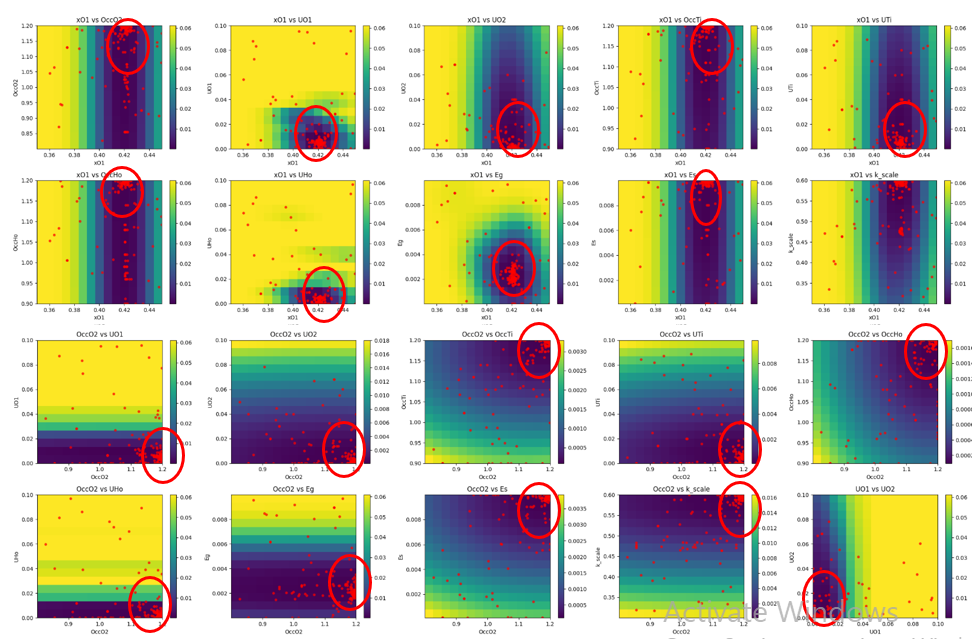}
    \caption{Figure showing the GP map variance for the TurBO implementation. As can be seen, the BO focuses more on the region of the parameter space where low GP variance is predicted. }
    \label{fig:Fig5-GP_Map_Var}
\end{figure}

\begin{figure}[H]
    \centering
    \includegraphics[width=0.9\linewidth]{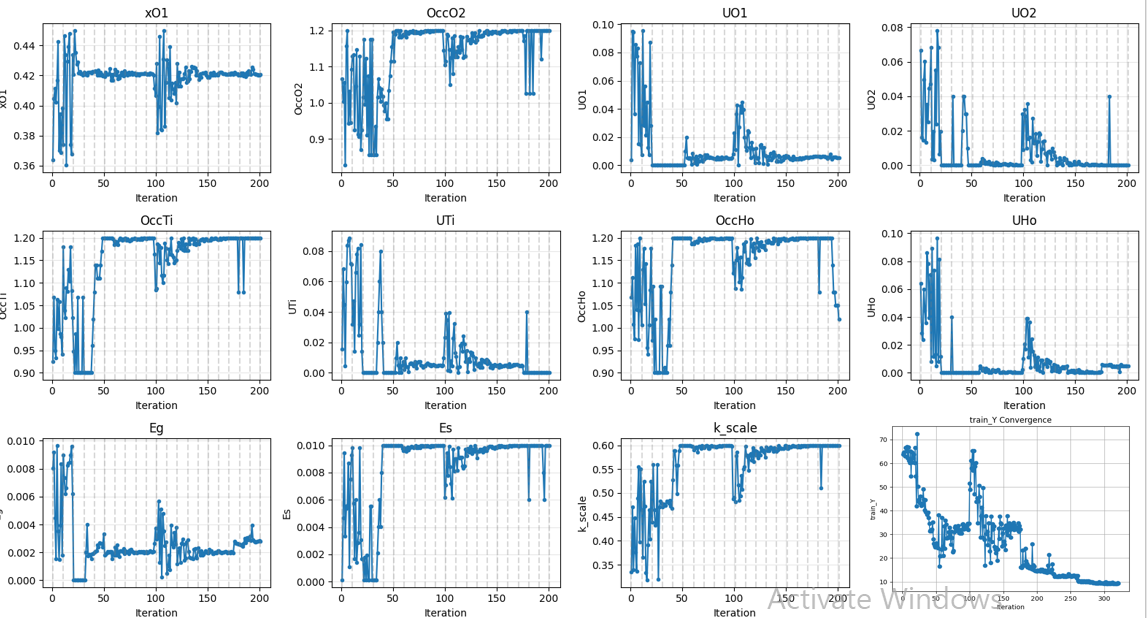}
    \caption{Figure showing the convergence of each of the parameters for the TuRBO workflow and the $\chi^2$ convergence curve for this BO workflow (the last figure on the bottom right corner }
    \label{fig:Fig6-Param-convergence}
\end{figure}

\section{X-ray Diffraction Experiment}
\label{sec:experiment}

We collected X-ray diffraction data from a high-quality single crystal of \(Ho_2Ti_2O_7\) grown using the traveling solvent floating zone (TSFZ) technique. The diffraction experiment was performed on a crystal of approximate dimensions \(0.1 \times 0.1 \times 0.1\ \text{mm}^3\) using a SuperNova diffractometer equipped with a Mo micro-focus sealed-tube X-ray source \((\lambda = 0.71073\ \text{\AA})\), a four-circle goniometer, and an Atlas CCD detector. Measurements were carried out at \(T = 110(2)\ \text{K}\) under a nitrogen atmosphere. Full reciprocal-space coverage was achieved through \(\omega\)-scan data collection. Diffraction intensities were recorded over the angular range \(3.5^\circ \leq \theta \leq 36.1^\circ\). Absorption corrections were applied using analytical and spherical-harmonics-based scaling procedures implemented in CrysAlisPro. The processed dataset was subsequently used for unit-cell refinement and structural analysis, yielding a generalized reflection file (\texttt{.hkl}) containing measured intensities \(I_{hkl}\) for reciprocal lattice vectors \( \mathbf{G} = h\mathbf{a}^* + k\mathbf{b}^* + l\mathbf{c}^* \), where \(\mathbf{a}^*, \mathbf{b}^*, \mathbf{c}^*\) are the reciprocal lattice basis vectors.

\section{Formulation of Refinement as BO Problem}
\label{sec:BO_form1}

Crystal structure refinement is an inverse problem in which model parameters are estimated by fitting a theoretical diffraction model to experimentally observed data. BO is routinely implemented using both X-ray and neutron scattering data. Let the observed diffraction intensities from either X-ray or neutron diffraction experiment be denoted as $I_{\text{obs}} = \{ I_{\text{obs},i} \}_{i=1}^{N}$, where $N$ is the number of measured reflections. The calculated intensity (model) for each reflection is given by
\begin{equation}
I_{\text{calc},i}(\theta, \vartheta) = k_{\text{scale}} \cdot L_i \cdot P_i \cdot T_i \cdot y_i(\vartheta) \cdot |F_{c,i}(\theta)|^2,
\label{eq:Icalc}
\end{equation} \cite{Rietveld1969}
where
\begin{equation}
F_{c,i}(\theta) = 
\sum_{j \in \{\mathrm{Ho},\mathrm{Ti},\mathrm{O}\}} 
x_j 
\sum_{n=1}^{N_j} 
\exp\left[2\pi i (h x_{jn} + k y_{jn} + l z_{jn})\right]
\exp\left(-B_j \frac{\sin^2\theta_i}{\lambda^2}\right)
\label{eqn:fcal}
\end{equation} \cite{ITCA2006, Cullity2001XRD, Giacovazzo2011Fundamentals}

Specifically for XRD and ND respectively, structure factors are:

\begin{equation}
F^{\mathrm{xrd}}_{c,i}(\theta)=
\sum_{j \in \{\mathrm{Ho},\mathrm{Ti},\mathrm{O}\}}
f_j(Q_i)
\sum_{n=1}^{N_j}
\exp\!\left[2\pi i(hx_{jn}+ky_{jn}+lz_{jn})\right]
\exp\!\left(-B_j\frac{\sin^2\theta_i}{\lambda^2}\right)
\label{eq:Fcal_xrd}
\end{equation}

\begin{equation}
F^{\mathrm{nd}}_{c,i}(\theta)=
\sum_{j \in \{\mathrm{Ho},\mathrm{Ti},\mathrm{O}\}}
b_j
\sum_{n=1}^{N_j}
\exp\!\left[2\pi i(hx_{jn}+ky_{jn}+lz_{jn})\right]
\exp\!\left(-B_j\frac{\sin^2\theta_i}{\lambda^2}\right)
\label{eq:Fcalc_nd}
\end{equation}

$k_{\text{scale}}$ is the scale factor, $L_i$ is the Lorentz factor, $P_i$ is the polarization factor, $T_i$ is the absorption/transmission factor, $y_i(\vartheta)$ is the extinction term, and $F_{c,i}(\theta)$ is the calculated structure factor. The structural parameter vector is $\theta = \{ \text{Occ}{\text{Ti}}, U{\text{Ti}}, \text{Occ}{\text{Ho}}, U{\text{Ho}}, x{\text{O}1}, \text{Occ}{\text{O}2}, U{\text{O}_1}, U{\text{O}_2} \}$, while $\vartheta = \{E_p, E_s, E_g\}$ denotes extinction-related parameters. A key distinction between the theoretical models for neutron and X-ray diffraction lies in the treatment of extinction effects. For \textbf{X-ray diffraction} model, extinction effects are typically negligible, and thus $y_i = 1$ \cite{Becker1974ExtinctionI, Becker1974ExtinctionII}. The model reduces to:
$I_{\text{calc},i}^{\text{X-ray}}(\theta) = k_{\text{scale}} \cdot L_i \cdot P_i \cdot T_i \cdot |F_{c,i}(\theta)|^2.$
In neutron diffraction model, the extinction term is defined as:
$y_i^{-1} = \sqrt{1 + C_G PF_i + \frac{A_G PF_i^2}{1 + B_G PF_i}}$, where $PF_i = PLZ_i \cdot \sigma_{\text{eff},i}$, $PLZ_i = A_V |F_{c,i}|^2 \lambda_i^2$, and $\sigma_{\text{eff},i} = \frac{E_g}{\sqrt{1 + (E_s PL_i / E_g)^2}}$ \cite{Becker1974ExtinctionI, Becker1974ExtinctionII}.

Thus, the neutron model is:
$I_{\text{calc},i}^{\text{neutron}}(\theta, \vartheta) = k_{\text{scale}} \cdot L_i \cdot P_i \cdot T_i \cdot y_i(\vartheta) \cdot |F_{c,i}(\theta)|^2.$

\textbf{Classical Optimization Objective}: Traditional refinement methods seek parameters that minimize the discrepancy between observed and calculated intensities. This is typically formulated as the weighted least-squares (chi-square) objective:

\begin{equation}
f(\theta, \vartheta) = \frac{1}{N}\sum_{i=1}^{N} \left( I_{\text{obs},i} - I_{\text{calc},i}(\theta, \vartheta) \right)^2,
\label{eq:chisq_w1}
\end{equation}

The refinement problem is therefore to solve:
\begin{equation}
(\theta^*, \vartheta^*) = \arg\min_{\theta, \vartheta} \chi^2(\theta, \vartheta).
\label{eq:opt}
\end{equation}

In practice, methods such as least squares and MLE solve Eq.~\ref{eq:opt} using local optimization, which can struggle with multimodality, parameter coupling, and noise.

\textbf{BO Formulation}: Instead of solving eq ~\ref{eq:opt} directly, we treat refinement as a black-box optimization problem over a bounded domain $\mathcal{D}$:
\begin{equation}
\min_{\theta, \vartheta \in \mathcal{D}} \ f(\theta, \vartheta).
\label{eq:bo_problem}
\end{equation}

\textbf{Initialization}: In the implementation of the BO frameworks, an initial design of $n$ parameter configurations $\{ (\theta_j, \vartheta_j) \}_{j=1}^{n}$ is generated using Latin Hypercube Sampling (LHS), ensuring space-filling coverage of $\mathcal{D}$. For each of the set of parameters, we calculate the corresponding $F_{\mathrm{cal},i}$ and objective values ($\chi^2(\theta, \vartheta)$) which is defined in this work as:

\begin{equation}
f(\theta, \vartheta) = \frac{1}{N} \sum_{i=1}^{N} \left( I_{\text{obs},i} - I_{\text{calc},i}(\theta, \vartheta) \right)^2,
\label{eq:chisq1}
\end{equation}

\textbf{BO Surrogate Model}: A Gaussian Process surrogate \cite{Rasmussen2006} is constructed as
\begin{equation}
f(\cdot) \sim \mathcal{GP}\big(\mu(\cdot), k(\cdot,\cdot)\big),
\label{eq:gp}
\end{equation}

where $\mu$ and $k$ denote the mean and covariance functions, respectively.

\textbf{BO Acquisition and Iteration}: At iteration $t$, the next evaluation point is selected via an acquisition function $\alpha(\cdot)$:
\begin{equation}
(\theta_{t+1}, \vartheta_{t+1}) = \arg\max_{\theta,\vartheta \in \mathcal{D}} \alpha(\theta, \vartheta \mid \mathcal{D}_t).
\label{eq:acq}
\end{equation}

For example, Expected Improvement (EI) is defined as $\alpha_{\text{EI}}(\theta) = \mathbb{E}[\max(0, \chi^2_{\text{best}} - \chi^2(\theta))]$.

The dataset is updated as $\mathcal{D}_{t+1} = \mathcal{D}_t \cup \{ (\theta_{t+1}, \vartheta_{t+1}, \chi^2(\theta_{t+1}, \vartheta_{t+1})) \}$, and the surrogate model is refit. This process is repeated until a predefined evaluation budget is reached \cite{Agada2025BORefinement}.

Unlike classical approaches that directly minimize Eq.~\ref{eq:chisq} using local search, Bayesian optimization constructs a global probabilistic model of the objective landscape and adaptively selects evaluation points. This enables efficient optimization in high-dimensional, non-convex, and noisy settings typical of crystal structure refinement in frustrated magnetic. Implementation was done with GPyTorch \cite{Gardner2018GPyTorch} and BoTorch \cite{Balandat2020BoTorch}.


\section{CrystalMO-TuRBO Implementation Details}
\subsection{Objectives and design variables}

Let $\bm{z}\in\mathcal{B}\subset\mathbb{R}^{d}$ denote a vector of $d$ tunable \emph{joint} set of parameters (structure, extinction, and scale parameters) with independent box bounds
\[
\mathcal{B} = \{\bm{z}\in\mathbb{R}^{d}:\; z_j^{\mathrm{lo}}\le z_j \le z_j^{\mathrm{hi}},\; j=1,\ldots,d\}.
\]
For each set of parameters $\bm{z}$, we evaluate the structure factors for $XRD$ and $ND$ using eq~\ref{eq:Fcal_xrd} and eg\ref{eq:Fcalc_nd}, and calculate intensities for each using eq~\ref{eq:Icalc}. Then, using the experimental data, the calculated intensities and evaluates two independent scalar \emph{mismatch} metrics that we take as objectives to \emph{minimize}. The \emph{objectives} are:
\[
f_1(\bm{z}) = f_{\mathrm{xrd}}(\bm{z}),\qquad
f_2(\bm{z}) = f_{\mathrm{nd}}(\bm{z}).
\]

It is numerically convenient to work in a globally normalized design space; hence, we define component-wise affines:
\[
x_j = \frac{z_j - z_j^{\mathrm{lo}}}{z_j^{\mathrm{hi}}-z_j^{\mathrm{lo}}}\in[0,1],
\qquad
\bm{x} = (x_1,\ldots,x_d)\in[0,1]^{d},
\]
(using a safe value when a range is zero), and let $\varphi:\mathcal{B}\to[0,1]^{d}$ be this map.

\subsection{Transformation to a Two-Objective Maximization Problem}
\label{sec:normalization}

The native refinement objectives $f_1(\bm{z})$ and $f_2(\bm{z})$ represent mismatch metrics to be minimized (e.g., X-ray and neutron experiment/model profile discrepancies). For compatibility with acquisition functions that are naturally posed as maximization problems, we transform the refinement task into a two-objective maximization problem. In addition, because the two objectives may differ substantially in scale, variance, or numerical conditioning, direct scalarization can bias optimization toward the larger-magnitude objective. To mitigate this, we normalize each objective using standardized residual scores prior to surrogate modeling and scalarization, a common practice in Gaussian process optimization and multi-objective Bayesian optimization \cite{Rasmussen2006,Snoek2012,Paria2020}.

Given an archive of $N$ evaluated parameter vectors 
$\{ \bm{z}^{(i)} \}_{i=1}^{N}$ with corresponding objective values 
$\{(f_1^{(i)},f_2^{(i)})\}_{i=1}^{N}$, we compute the empirical mean and standard deviation of each objective:

\[
\bar f_m = \frac{1}{N}\sum_{i=1}^{N} f_m^{(i)}, \qquad
s_m = \max\!\left\{\varepsilon,\sqrt{\frac{1}{N-1}\sum_{i=1}^{N}(f_m^{(i)}-\bar f_m)^2}\right\},
\quad m\in\{1,2\},
\]

where $\varepsilon>0$ is a small numerical floor introduced to prevent division by zero when variance is negligible.

We then define transformed objectives:

\begin{equation}
Y_m^{(i)} = -\frac{f_m^{(i)}-\bar f_m}{s_m},
\qquad m\in\{1,2\},
\label{eq:Yscalar}
\end{equation}

and stack them into the reward vector

\begin{equation}
\bm{Y}^{(i)} =
\begin{pmatrix}
Y_1^{(i)} \\
Y_2^{(i)}
\end{pmatrix}.
\label{eq:Yvector}
\end{equation}

This transformation has three desirable properties. First, centering removes absolute scale dependence and improves numerical conditioning of Gaussian process regression \cite{Rasmussen2006}. Second, variance normalization places both objectives on comparable magnitude, preventing domination of one objective during scalarization \cite{Paria2020}. Third, the negative sign converts the original minimization problem into a reward-maximization problem, so lower diffraction mismatch corresponds to larger objective values.

The statistics $(\bar f_m,s_m)$ are recomputed from the current archive after each optimization update. Consequently, improvements in either original refinement objective induce corresponding increases in at least one component of $\bm{Y}$, enabling adaptive and balanced multi-objective search throughout the optimization process.

\subsection{Phase~1: Multi-Objective TuRBO via Parallel Scalarized Trust Regions}
\label{subsec:phase1-appendix}

Phase~1 is a new extension of trust-region Bayesian optimization that adapts TuRBO \cite{Eriksson2019TuRBO} to the multi-objective joint refinement setting through parallel scalarized sub-problems. Rather than directly optimizing two competing objectives simultaneously, we decompose the problem into $K$ independent scalarized Bayesian optimization tasks, each associated with a different trade-off between X-ray and neutron refinement quality. This combines ideas from trust-region BO \cite{Eriksson2019TuRBO}, scalarization-based multi-objective optimization \cite{Paria2020,Miettinen1999}, and batch Expected Improvement \cite{Jones1998EGO,Balandat2020BoTorch}. We adopt scalarized q-Expected Improvement within parallel trust regions rather than Expected Hypervolume Improvement (EHVI) because scalarization-based acquisition offers lower computational overhead, superior scalability in high-dimensional parameter spaces, and natural compatibility with TuRBO-style local search. These advantages are especially important for expensive crystal refinement objectives involving many coupled structural parameters \cite{Eriksson2019TuRBO,Paria2020,Daulton2020qEHVI}. While linear scalarization fixes a single direction in objective space and can therefore under-represent parts of the Pareto front that are poorly aligned with that weight, a well-known limitation of weighted-sum formulations, our procedure still records both objectives at every evaluation, so the empirical trade-off set remains available for a posteriori Pareto assessment and visualization \cite{Miettinen1999}

\textbf{Scalarized sub-objectives}: For each trust-region state $k=1,\dots,K$, we define a scalarized maximization objective using convex combinations of the transformed objectives:
\begin{equation}
S_k(\bm{Y}) = \lambda_k Y_1 + (1-\lambda_k)Y_2,
\label{eq:Sk}
\end{equation}
where $\lambda_k \in (0,1)$ specifies the preference weight for objective $Y_1$, while $1-\lambda_k$ weights $Y_2$. Different values of $\lambda_k$ allow simultaneous exploration of different trade-offs along the Pareto frontier \cite{Paria2020}.

\textbf{Parallel trust-region states}: Each scalarized objective is assigned an independent TuRBO--1 state \cite{Eriksson2019TuRBO}. The center of trust region $k$ is chosen as the best previously observed solution under scalarization $S_k$:

\begin{equation}
\bm{x}^{(k)}_{\mathrm{c}} \in 
\arg\max_{\bm{x}^{(i)} \in \mathcal{A}}
S_k(\bm{Y}^{(i)}),
\label{eq:center}
\end{equation}
where $\mathcal{A}$ denotes the global archive of evaluated solutions. Around this center, a hyper-rectangular trust region $\mathcal{R}_k \subset [0,1]^d$ is maintained and adaptively expanded or contracted using the standard TuRBO success/failure rules \cite{Eriksson2019TuRBO}.

\textbf{Surrogate modeling}: For each region $k$, we fit an independent Gaussian Process surrogate model:


$S_k(\bm{x}) \sim \mathrm{GP}\left(\mu_k(\bm{x}), k_k(\bm{x},\bm{x}')\right)$\cite{Rasmussen2006}

where $\mu_k(\cdot)$ and $k_k(\cdot,\cdot)$ denote the posterior mean and covariance functions.
\begin{figure}[t]
    \centering
    \includegraphics[width=0.9\linewidth]{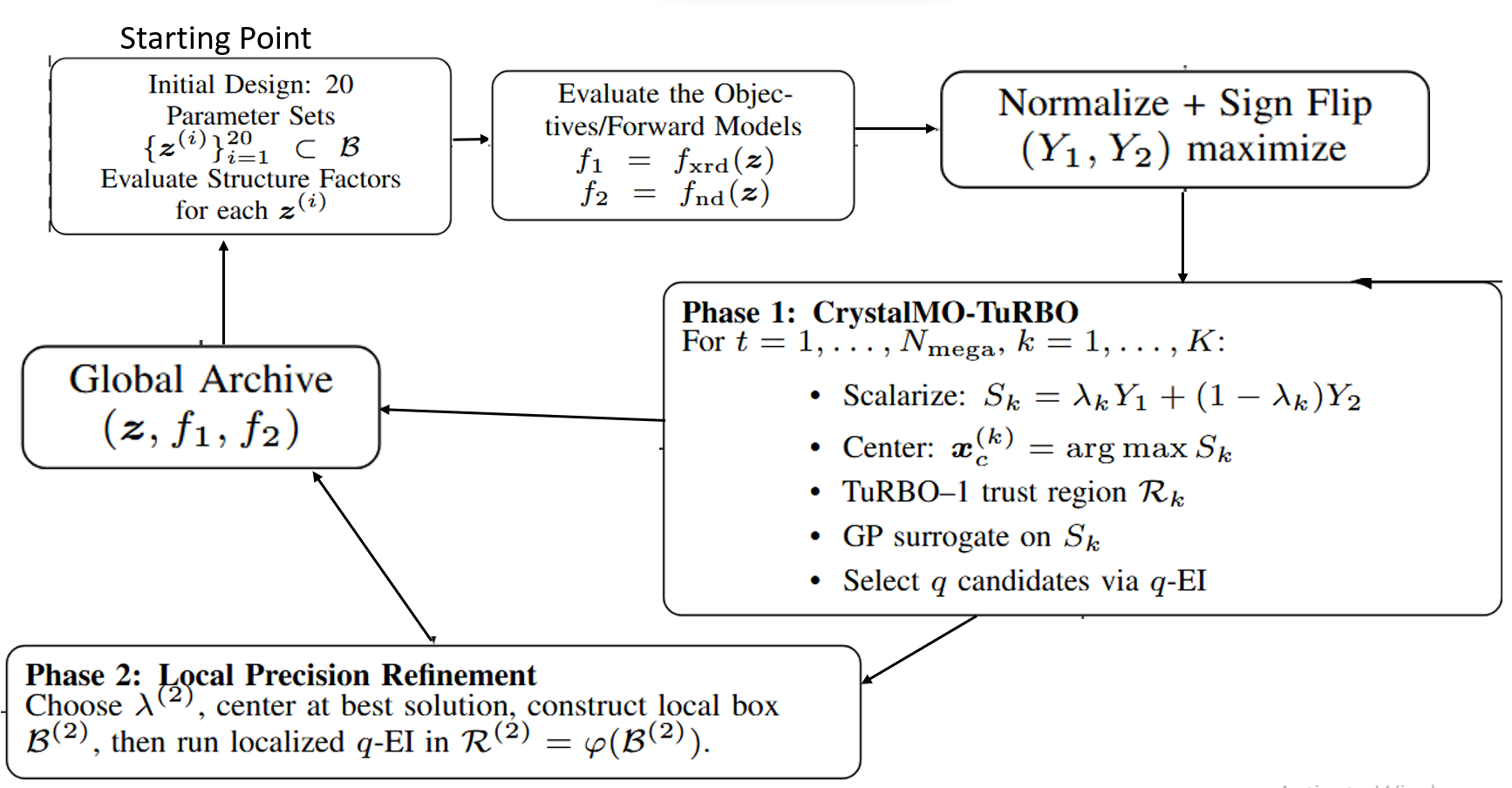}
    \caption{This Figure visually depicts the proposed \textit{CrytalMO-TuRBO} workflow. From the starting point, we select $n=20$ ($n$ is a hyperparameter that can be tuned) initial sets of parameters (design variables), and we then calculate the structure factor for each of the sets of parameters, for each of the X-ray and neutron diffraction data. At the next stage, using the experimental X-ray and neutron diffraction data, with the evaluated structure factors, we calculate the X-ray and neutron objective values for each of the initial set of variables. At the next stage, we normalize the objectives and flip the sign to convert the problem to a maximization problem. Then we move on to implement the phase 1 of the proposed CrystalMO-TuRBO framework as described in Section~\ref{subsec:phase1}. The new solution from this stage is added to the Global Archive. The solution(s) in the Global Archive is/are added to the initial design variables, and this is repeated until Phase 1 budget is exhausted. Once Phase 1 budget is xhauted, we skip phase in for each iteration, select the next solution at each stage using Phase 2 as described in Section~\ref{sec:phase_two}}
    \label{fig:Fig7-CrystalMO-TuRBO}
\end{figure}

\textbf{Candidate generation via qEI}: Within each trust region, a batch of $q$ candidate solutions is selected by maximizing the $q$-Expected Improvement acquisition function \cite{Jones1998EGO,Balandat2020BoTorch} and the next solution for each trust region is chosen as:
\begin{equation}
\bm{x}_{\text{next}}^{(k)} =
\arg\max_{\bm{x}\in\mathcal{R}_k}
\alpha_{\text{qEI}}(\bm{x}).
\label{eq:qei_phase1}
\end{equation}
Each proposed point is mapped back to the physical parameter space, jointly evaluated on X-ray and neutron diffraction objectives, and appended to the archive. This cycle continues until phase 1 budget is exhausted.

\textbf{Global iterative loop:} After initialization using Latin Hypercube Sampling \cite{McKay1979LHS}, Phase~1 performs $N_{\mathrm{mega}}$ outer sweeps. In each sweep, all $K$ trust regions generate new candidates and update their states. This produces at most $K \times N_{\mathrm{mega}}$ new evaluations, excluding initialization points.

\subsection{Phase 2: TURBO--style local q--EI in a fixed axis-aligned box (single scalar)}
\label{sec:phase_two_appendix}

When Phase~1 budget is exhausted, a second phase continues to refine the search \emph{locally} in \emph{physical} space around the current best with respect to a \emph{single} user-chosen weight $\lambda^{(2)}$:
\begin{equation}
\label{eq:Sphase2}
S^{(2)}(\bm{Y}) = \lambda^{(2)} Y_1 + \bigl(1-\lambda^{(2)}\bigr)Y_2.
\end{equation}
Let the reference center $\bm{z}_{\ast}$ be an archive point that maximizes $S^{(2)}$ on the current dataset.

\paragraph{Local box:} From $\bm{z}_{\ast}$ and the global bounds, we form an \emph{axis-aligned} local box
\(
\mathcal{B}^{(2)} \subset \mathcal{B}
\)
as a fixed fraction of the global range in each dimension (in our implementation, a percentage of the distance from the best component to a bound; if a component of $\bm{z}_{\ast}$ is near~0, a fallback based on the global half-range is used.

\textbf{Local $q$--EI in normalized box:} The local box is mapped to normalized bounds $\mathcal{R}^{(2)}=\varphi\bigl(\mathcal{B}^{(2)}\bigr) \subset[0,1]^{d}$. For each new candidate we fit the same kind of one-dimensional score
\(
S = \lambda^{(2)} Y_1 + (1-\lambda^{(2)})Y_2
\)
on the full archive, build a \emph{scalar} GP in $\bm{x}$--space, and run \texttt{$q$--EI} \textbf{constrained to $\mathcal{R}^{(2)}$ (i.e. $q$--EI} in a fixed axis-aligned sub-box in normalized coordinates.

A loop tracks improvement of $S^{(2)}$; optional stopping rules cap the number of local trials. This phase uses independent hyperparameters $(\text{restarts}, \text{raw samples}, \text{box width\,\%}, \text{max evals per phase})$.

\subsection{Procedure for Selecting the Best-Compromise Solution}
\label{sec:utopia-compromise}

Multi-objective optimization produces a set of non-dominated (Pareto-optimal) solutions rather than a single optimum. In joint X-ray--neutron refinement, some solutions favor XRD agreement, whereas others favor ND agreement. Although the complete Pareto front is valuable for understanding the trade-off between the two modalities, practical refinement requires a single representative parameter vector for reporting structural parameters and comparison with single-solution methods. We therefore select a \emph{best-compromise} solution based on its proximity to the ideal (utopia) point.

\paragraph{Ideal (utopia) point.}
Let $\mathcal{P}$ denote the empirical Pareto set with objective vectors
\[
\left\{\left(f_{\mathrm{XRD}}^{(j)},f_{\mathrm{ND}}^{(j)}\right):j\in\mathcal{P}\right\}.
\]
The ideal point is defined as the component-wise minimum objective values on the Pareto front,

\begin{equation}
\mathbf{f}^{\mathrm{ut}}
=
(u_{\mathrm{XRD}},u_{\mathrm{ND}})^\top
=
\left(
\min_{j\in\mathcal{P}}f_{\mathrm{XRD}}^{(j)},
\;
\min_{j\in\mathcal{P}}f_{\mathrm{ND}}^{(j)}
\right)^\top.
\label{eq:utopia}
\end{equation}

Although generally unattainable, this point provides a natural reference for measuring the quality of Pareto-optimal solutions.

\paragraph{Normalized distance to the ideal point.}
Because the XRD and ND objectives may have different numerical scales, each objective is normalized using its range over the Pareto front,

\begin{equation}
r_{\mathrm{XRD}}
=
\max_{j\in\mathcal{P}}f_{\mathrm{XRD}}^{(j)}
-
\min_{j\in\mathcal{P}}f_{\mathrm{XRD}}^{(j)},
\qquad
r_{\mathrm{ND}}
=
\max_{j\in\mathcal{P}}f_{\mathrm{ND}}^{(j)}
-
\min_{j\in\mathcal{P}}f_{\mathrm{ND}}^{(j)}.
\label{eq:ranges}
\end{equation}

For numerical stability, any range smaller than $10^{-15}$ is replaced by $10^{-15}$. The normalized Euclidean distance of Pareto solution $j$ from the ideal point is then

\begin{equation}
d_j=
\sqrt{
\left(
\frac{f_{\mathrm{XRD}}^{(j)}-u_{\mathrm{XRD}}}{r_{\mathrm{XRD}}}
\right)^2
+
\left(
\frac{f_{\mathrm{ND}}^{(j)}-u_{\mathrm{ND}}}{r_{\mathrm{ND}}}
\right)^2 }.
\label{eq:compromise_distance}
\end{equation}

The reported solution is the Pareto-optimal point with the smallest normalized distance,

\begin{equation}
j^\star
=
\arg\min_{j\in\mathcal{P}} d_j.
\label{eq:best_compromise}
\end{equation}

This criterion selects the Pareto solution that most evenly balances the XRD and ND objectives without introducing user-defined weights during optimization. Unlike weighted-sum scalarization, the compromise rule is applied only after the Pareto front has been constructed and therefore does not bias the search trajectory. It provides a transparent, scale-invariant, and reproducible procedure for selecting a single representative solution that can be directly compared with conventional least-squares refinement and other single-solution optimization methods.

\section{MORBO Method}

We implemented \emph{Multi-Objective Bayesian Optimization with Random Scalarizations}
(MORBO)~\cite{Daulton2022MORBO}: several trust regions (TRs) explore the
parameter space in parallel on a \emph{shared} surrogate, while acquisition
uses \emph{random convex scalarizations} of objective vectors together with
\emph{Thompson sampling} (TS). TR geometry and length adaptation follow
TuRBO-style rules keyed to multi-objective improvement via hypervolume.

\paragraph{Problem setup.}
Let $z \in \mathcal{Z} \subset \mathbb{R}^d$ denote the decision vector (here,
joint crystal / instrument parameters). Two expensive black-box objectives are
computed (e.g.\ XRD and ND $\chi^2$), both to be \emph{minimized}. For use inside
MORBO (which fits GP models on a \emph{maximization} convention), the notebook
maps raw losses $(r_{\mathrm{xrd}}, r_{\mathrm{nd}})$ to \emph{maximization}
targets $y^{(1)}, y^{(2)}$ by standardizing each raw stream to zero mean and
unit scale on the observed training data, then negating (higher is better).

\paragraph{Surrogate model.}
All evaluated points $(z_i, y_i)$ with $y_i \in \mathbb{R}^2$ are used to fit
a \texttt{ModelListGP}: two independent single-task Gaussian processes, one
per objective, on inputs mapped to $[0,1]^d$ by componentwise affine scaling
from box bounds (``01 normalization'').

\paragraph{Trust-region initialization.}
With $K$ trust regions, centers are chosen from the current nondominated set in
\emph{objective} space when possible: a greedy \emph{maximin} heuristic in
\emph{input} space spreads the $K$ centers among nondominated 01-points. If fewer
than $K$ nondominated points exist, centers are drawn from all data or padded
by repetition.

\paragraph{Proposal mechanism (one new evaluation per iteration).}
For each TR $k$:
\begin{enumerate}
  \item Define a TR as a hypercube in 01-space centered at $c_k$ with edge
        length $\ell_k$ (clipped to $[0,1]^d$).
  \item Identify training points whose 01-coordinates lie inside this hypercube.
        Among them, restrict to points that are nondominated \emph{with respect
        to the objectives observed inside that TR}; if none, fall back to the
        center $c_k$. Call this finite set of ``local incumbent'' inputs
        $\mathcal{X}_k^{\mathrm{loc}}$.
  \item Build a large discrete candidate pool inside the TR by TuRBO-style
        \emph{subset perturbation} of local incumbents: with probability
        decaying as the evaluation budget grows, each candidate perturbs a random
        subset of coordinates using either Sobol (QMC) or uniform draws inside
        the TR bounds.
  \item Draw a single vector $\lambda$ from the 2-simplex (uniformly over
        $(\lambda_1,\lambda_2)$ with $\lambda_j \ge 0$, $\lambda_1+\lambda_2=1$).
  \item Draw one joint \emph{Thompson} fantasy $\tilde{f}(x)$ from the
        \texttt{ModelListGP} posterior at all candidates $x$ in the pool, and
        rank candidates by the \emph{scalarized} score
        $\lambda_1 \tilde{f}_1(x) + \lambda_2 \tilde{f}_2(x)$.
  \item Select the best candidate within TR $k$.
\end{enumerate}
We repeat the above for \emph{all} $K$ TRs in one iteration and evaluate the
single point with the largest scalarized TS value across regions (MORBO's greedy
multi-TR step).

\paragraph{Hypervolume and TR updates.}
Let $Y$ be the matrix of observed maximization targets after the new evaluation.
We fix a reference point $r$ slightly below the per-dimension minima of the
current data (5\% margin on each objective range) and compute dominated
hypervolume of the \emph{nondominated} subset of $Y$ above $r$ using BoTorch's
\texttt{DominatedPartitioning}. Comparing hypervolume before vs.\ after the
new point flags improvement. For the TR that proposed the new point:
\begin{itemize}
  \item On improvement: increment a success counter; optionally recenter the TR
        on the nondominated point inside the TR whose \emph{sum}
        $(y^{(1)} + y^{(2)})$ is largest (a simple scalar tie-break among local
        Pareto points); double the TR length after enough consecutive successes
        (capped at $\ell_{\max}$).
  \item On no improvement: increment a failure counter; halve the TR length after
        enough consecutive failures (floored at $\ell_{\min}$). If the length
        hits the floor, reseed the center by picking a random globally
        nondominated archived point and reset the length to $\ell_{\mathrm{init}}$.
\end{itemize}

\paragraph{Relation to the reference.}
This matches the MORBO recipe of Daulton et al.: coordinated multi--trust-region
search, shared multi-output uncertainty via \texttt{ModelListGP}, TS-based
acquisition, and random scalarization weights for scalar ranking~\cite{Daulton2022MORBO}.
Our file \texttt{morbo\_joint.py} is a compact BoTorch-only re-implementation
(not the original Meta repository).

\paragraph{Implementation note.}
Diagnostics (Pareto plots, GP slice visualizations, etc.) are produced from the
archived $(z_i, r_{\mathrm{xrd},i}, r_{\mathrm{nd},i})$ and the normalized
targets used inside MORBO; they do not change the optimization dynamics.

\section{ParEGO Method}
ParEGO~\cite{Knowles2006ParEGO} is a sequential surrogate-based method for
multi-objective optimization when each objective evaluation is expensive. Its
central idea is to reduce the vector-valued problem to a \emph{sequence} of
single-objective problems: at each iteration one draws a new random weight
vector on the simplex and optimizes an \emph{augmented Chebyshev} (Tchebycheff)
scalarization of the objectives. A single Gaussian process (GP) is fit to the
resulting scalar ``cost'' on all past evaluations, and the next point is chosen
by maximizing expected improvement (EI) [or a similar one-step lookahead] on
that scalar surrogate. Changing the weights across iterations drives exploration
of different trade-offs and, in the limit, supports filling out the Pareto front.

\paragraph{Problem setup.}
Let $f : \mathcal{X} \to \mathbb{R}^m$ be a vector of objectives to
\emph{minimize}, with $\mathcal{X} \subset \mathbb{R}^d$ bounded (box constraints
in practice). Denote $f(x) = (f_1(x),\ldots,f_m(x))^\top$.

\paragraph{Ideal / utopia point.}
From the archive of evaluations $\{f(x^{(n)})\}_{n=1}^N$, form a \emph{utopia}
(ideal) reference $z^\star \in \mathbb{R}^m$ by taking, for each coordinate $j$,
a value strictly better than the best $f_j$ observed so far (e.g.\ the empirical
minimum minus a small margin). This stabilizes the scalarization and mimics
``aspiration levels'' lying beyond the current best per objective.

\paragraph{Augmented Chebyshev scalarization.}
At iteration $t$, sample a weight vector $\lambda = (\lambda_1,\ldots,\lambda_m)$
with $\lambda_j \ge 0$ and $\sum_j \lambda_j = 1$ (uniform on the simplex is the
usual choice). Define the scalar objective
\begin{equation}
  s_\lambda(x)
  \;=\;
  \max_{j \in \{1,\ldots,m\}} \lambda_j \,\bigl(f_j(x) - z^\star_j\bigr)
  \;+\;
  \rho \sum_{j=1}^m \bigl(f_j(x) - z^\star_j\bigr),
  \label{eq:parego_scalar}
\end{equation}
with small $\rho > 0$. The \emph{max} term is the (weighted) Chebyshev part; it
promotes movement toward compromise solutions aligned with $\lambda$. The
\emph{sum} term augments the scalarization so that improving one objective
without worsening others can still decrease $s_\lambda(x)$---mitigating
weaknesses of pure Chebyshev scalars along flats of the Pareto surface and
helping numerical optimization in practice.

\paragraph{Surrogate and acquisition.}
For all previously evaluated points $\{x^{(i)}\}_{i=1}^N$, compute scalars
$t_i = s_\lambda(x^{(i)})$. Fit a single GP regression model to
$\{(x^{(i)}, t_i)\}$, treating $t_i$ as a stationary, spatially correlated
expensive function of $x$ (standard GP + marginal likelihood or MAP hyperparameters).
Let $\mu(x)$ and $\sigma(x)$ be the GP predictive mean and standard deviation.
Select
\begin{equation}
  x^{(N+1)} \in \arg\max_{x \in \mathcal{X}} \mathrm{EI}(x),
\end{equation}
where $\mathrm{EI}$ is expected improvement over the best scalar value observed
so far for \emph{this} $\lambda$, with plug-in GP moments (possibly with
constraint handling via penalties or constrained EI variants if constraints are
present). After observing $f(x^{(N+1)})$, append to the archive, update
$z^\star$ if appropriate, and proceed to a new iteration with a \emph{fresh}
random $\lambda$.

\paragraph{Interpretation.}
Unlike methods that maintain explicit Pareto-aware acquisition on the vector
objective (e.g.\ hypervolume or expected R2 indicators), ParEGO navigates the
multi-objective landscape indirectly: each iteration solves a \emph{different}
weighted Chebyshev program under model uncertainty. Randomized scalarizations
approximate a diffuse search for the entire efficient set; the GP on $s_\lambda$
acts as a cheap global trend surface for that scalarized landscape.

\paragraph{Relation to other MOBO lines.}
ParEGO is an early, conceptually simple \emph{scalarization-first} approach.
Random scalarizations also appear inside other algorithms (e.g.\ some TS /
draw-based MOBO variants), but ParEGO is distinctive in pairing \emph{random
Chebyshev weights each step} with \emph{one GP on the scalarized objective} and
\emph{EI in the scalarized space}. Trust-region or multi-region extensions (such
as MORBO-style search) are not part of the original ParEGO recipe.

\section{qEHVI MOBO Method}

Expected Hypervolume Improvement (EHVI) \cite{Emmerich2006EHVI, Hupkens2014EHVI}, sometimes written as ``eEHVI'' when
emphasizing the \emph{expectation} under uncertainty, is a one-step lookahead
criterion for multi-objective Bayesian optimization (MOBO). At each
iteration, a probabilistic model (typically independent GPs or a multi-output GP
for the objectives) provides predictive beliefs about vector outcomes
$f(x)\in\mathbb{R}^m$. The next evaluation maximizes the \emph{expected
increase} of dominated hypervolume relative to a fixed \emph{reference point}
and the current empirical Pareto front. Unlike ParEGO-style random
scalarizations, EHVI is \emph{directly} aligned with hypervolume as a set-quality
indicator and therefore targets the compromise structure of the efficient set.

\paragraph{Problem setup and dominance.}
Let $f:\mathcal{X}\to\mathbb{R}^m$ be $m$ expensive objectives to
\emph{minimize} on a compact domain $\mathcal{X}\subset\mathbb{R}^d$. For two
points $u,v\in\mathbb{R}^m$, $u$ dominates $v$ (written $u\prec v$) if
$u_j\le v_j$ for all $j$ with at least one strict inequality. The optima form the
Pareto set in $\mathcal{X}$ and the Pareto front in objective space.

\paragraph{Hypervolume indicator and improvement.}
Fix a \emph{reference point} $r\in\mathbb{R}^m$ strictly worse than all
objectives of interest (componentwise: $r_j > \max_{x\in\mathcal{X}} f_j(x)$ in
minimization), so that hypervolume is well-defined and finite. For a finite set
of objective vectors $A=\{y^{(1)},\ldots,y^{(N)}\}$, the hypervolume of the
region dominated by $A$ and bounded below by $r$ is
\begin{equation}
  \mathrm{HV}(A; r)
  \;=\;
  \Lambda\Bigl(\,\bigcup_{i=1}^N \bigl\{ y\in\mathbb{R}^m : r \prec y \prec y^{(i)} \bigr\}\Bigr),
\end{equation}
where $\Lambda$ is Lebesgue measure (volume). Given current nondominated
observations, let $P_N$ denote their objective vectors after removing dominated
points; $\mathrm{HV}_N := \mathrm{HV}(P_N; r)$ measures multi-objective
progress. If evaluating $x$ yields $Y=f(x)\in\mathbb{R}^m$, the (random)
\emph{hypervolume improvement} is
\begin{equation}
  \mathrm{HVI}(Y)
  \;=\;
  \max\bigl\{0,\ \mathrm{HV}(P_N \cup \{Y\}; r) - \mathrm{HV}_N \bigr\},
\end{equation}
which is zero if $Y$ is dominated by the current archive contribution.

\paragraph{EHVI acquisition (``eEHVI'').}
A vector surrogate provides a predictive distribution $Y\mid x, \mathcal{D}_N$
(e.g.\ independent Gaussian marginals from $m$ GPs, or a correlated model).
The EHVI acquisition is
\begin{equation}
  \alpha_{\mathrm{EHAI}}(x)
  \;=\;
  \mathbb{E}_{Y\mid x,\mathcal{D}_N}\bigl[\,\mathrm{HVI}(Y)\,\bigr],
\end{equation}
the expected nonnegative volume increment added by one observation at $x$.
The next design is
\begin{equation}
  x_{N+1} \in \arg\max_{x\in\mathcal{X}} \alpha_{\mathrm{EHAI}}(x),
\end{equation}
possibly adjusted for constraints or batches.

\paragraph{Computation and modeling notes.}
$\mathrm{HVI}(Y)$ is a piecewise function of $Y$ determined by dominated
partitions relative to $(P_N,r)$; the expectation integrates against the
predictive density of $Y\mid x$. For $m=2$, fast or semi-analytic
decompositions are common; for $m\ge 3$, implementations often rely on box
decompositions of dominated hypervolume (cell unions) combined with numerical
integration or Monte Carlo over the predictive distribution, especially for
correlated objectives. Software stacks (e.g.\ BoTorch utilities around dominated
hypervolume bookkeeping) mirror these constructions for $m=2$ and $m>2$.

\paragraph{Batch and noisy variants (context).}
The single-point EHVI principle extends to \emph{parallel} $\mathrm{q}$EHVI by
evaluating expected hypervolume improvement from $q$ pending points under a
joint predictive model (possibly with fantasies). Noisy observations require
defining hypervolume with respect to \emph{posterior} beliefs about objectives
or filtered nondominated sets; acquisition definitions follow the same ``expected
gain in hypervolume'' logic but with more intricate statistics.

\paragraph{Relation to scalarization \& MORBO.}
ParEGO explores the efficient set indirectly via random Chebyshev scalars and
one scalarized GP per iteration. MORBO combines multi-trust-region search with
TS/scalarization-based ranking. EHVI MOBO instead optimizes a \emph{single}
Pareto-aware functional---expected dominated volume gain---and is a standard
reference in the indicator-based MOBO line.






\section{Instruction for Code Implementation}
\label{code_implementation}
The code and data for this project are hosted on Open Science Framework with the following anonymized link: \url{https://osf.io/4n95u/overview?view_only=a7a8e5a8985442628c91185936076217}
All the .ipynb notebooks for each implementation are provided. To run the notebooks, place the .hkl data files for X-ray and neutron diffraction data, and the .py scripts provided in the same folder as the notebook, and run.  The script and .hkl files are in the home directory of the zipped project repository. Please ensure the setup of an appropriate environment

\section{Additional Plots}

\begin{figure}[t]
    \centering
    \includegraphics[width=\linewidth]{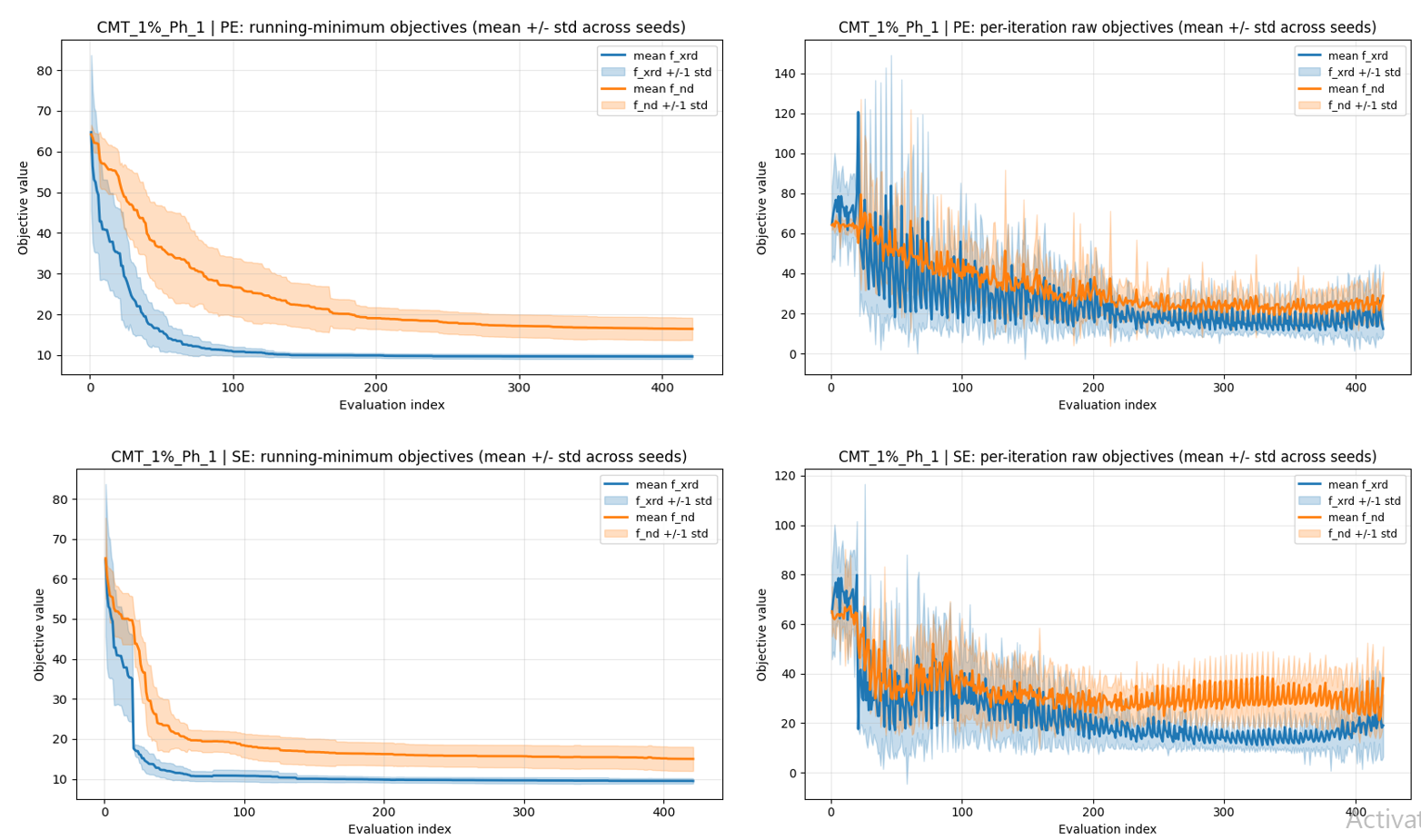}
    \caption{Figures showing convergence curve for the multiple run of the experiments for CMT Pase 1 only PE and SE 1\% bounding box width. Values plotted are means and standard deviations of objective values. Column 1 and 2 for each methods are running minimumns and raw objective values respectively.}
    \label{fig:appendix_one}
    \vspace{-2mm}
\end{figure}

\begin{figure}[t]
    \centering
    \includegraphics[width=\linewidth]{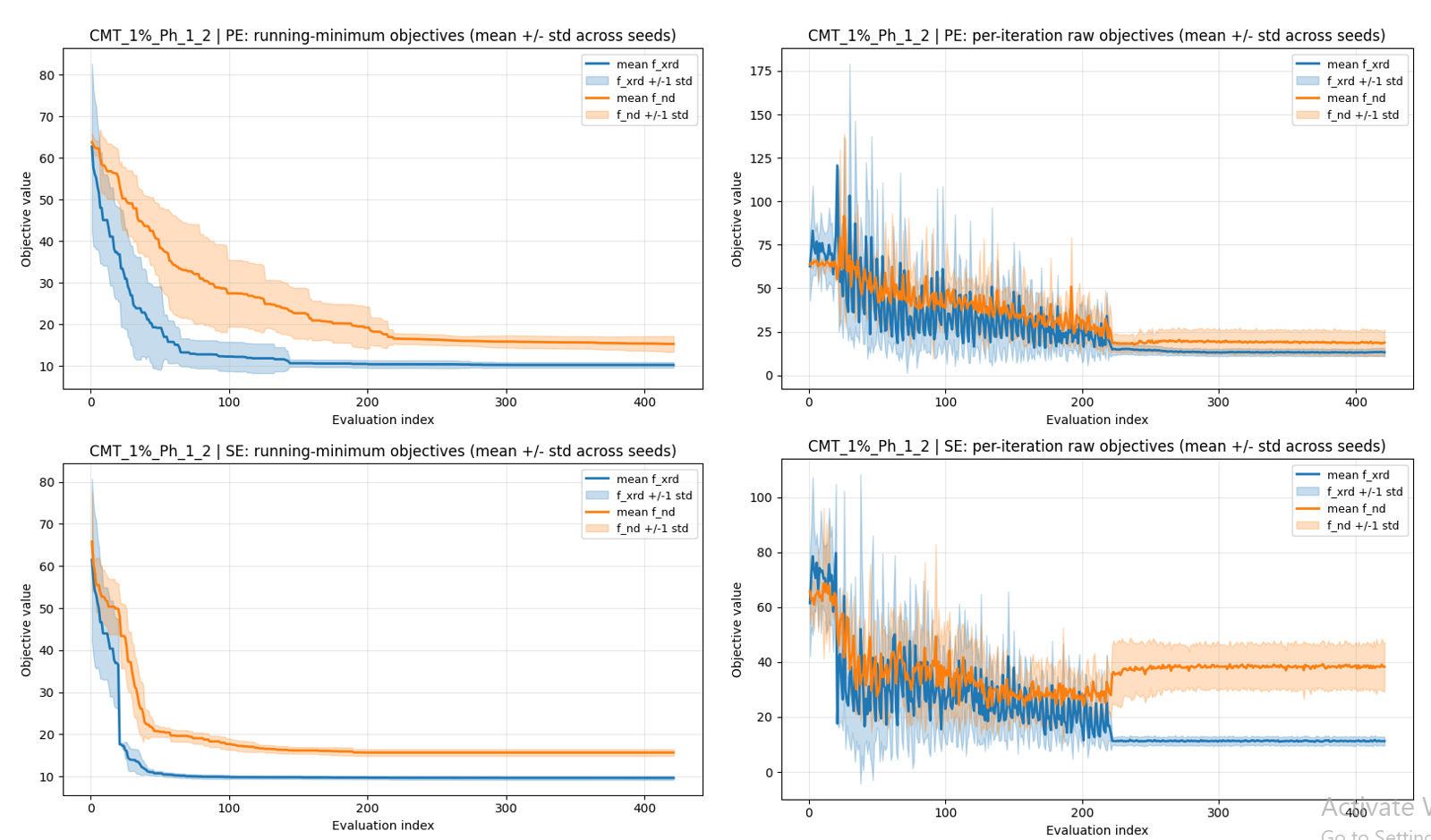}
    \caption{Figures showing convergence curve for the multiple run of the experiments for CMT Pase 1 and 2 PE and SE 1\% bounding box width. Values plotted are means and standard deviations of objective values. Column 1 and 2 for each methods are running minimumns and raw objective values respectively.}
    \label{fig:appendix_two}
    \vspace{-2mm}
\end{figure}

\begin{figure}[t]
    \centering
    \includegraphics[width=\linewidth]{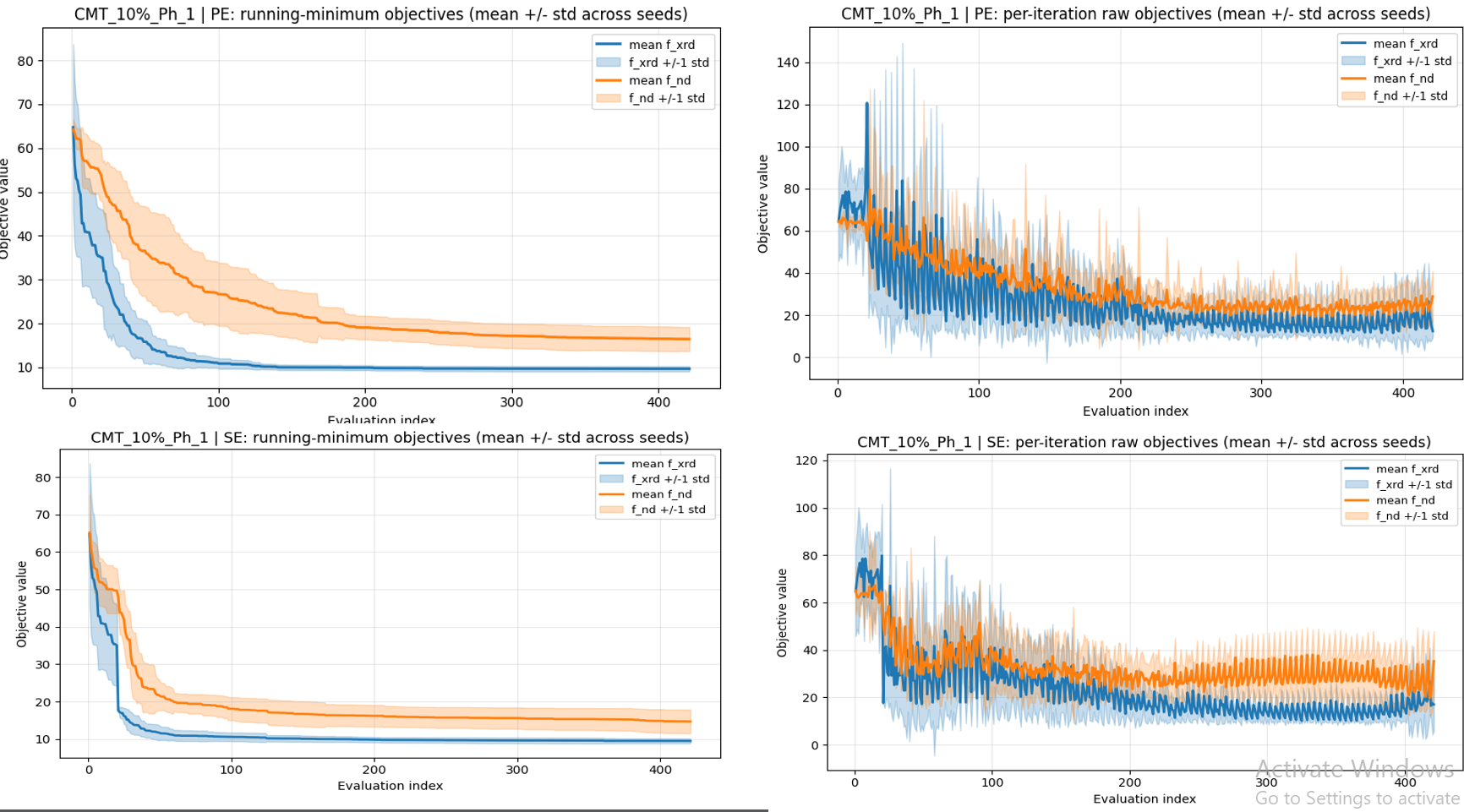}
    \caption{Figures showing convergence curve for the multiple run of the experiments for CMT Pase 1 only PE and SE 10\% bounding box width. Values plotted are means and standard deviations of objective values. Column 1 and 2 for each methods are running minimumns and raw objective values respectively.}
    \label{fig:appendix_three}
    \vspace{-2mm}
\end{figure}

\begin{figure}[t]
    \centering
    \includegraphics[width=\linewidth]{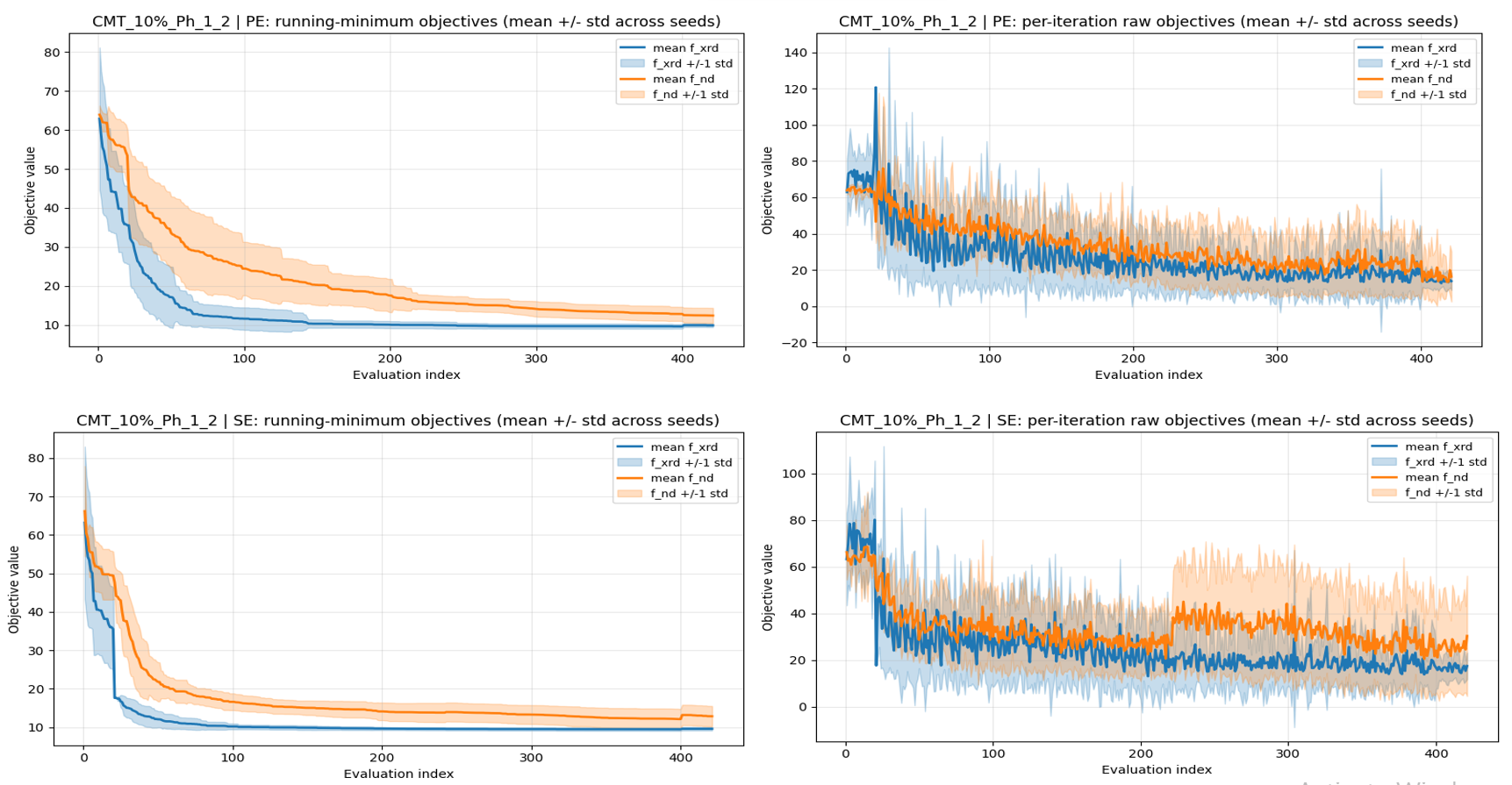}
    \caption{Figures showing convergence curve for the multiple run of the experiments for CMT Pase 1 and 2 PE and SE 10\% bounding box width. Values plotted are means and standard deviations of objective values. Column 1 and 2 for each methods are running minimumns and raw objective values respectively.}
    \label{fig:appendix_four}
    \vspace{-2mm}
\end{figure}

\begin{figure}[t]
    \centering
    \includegraphics[width=\linewidth]{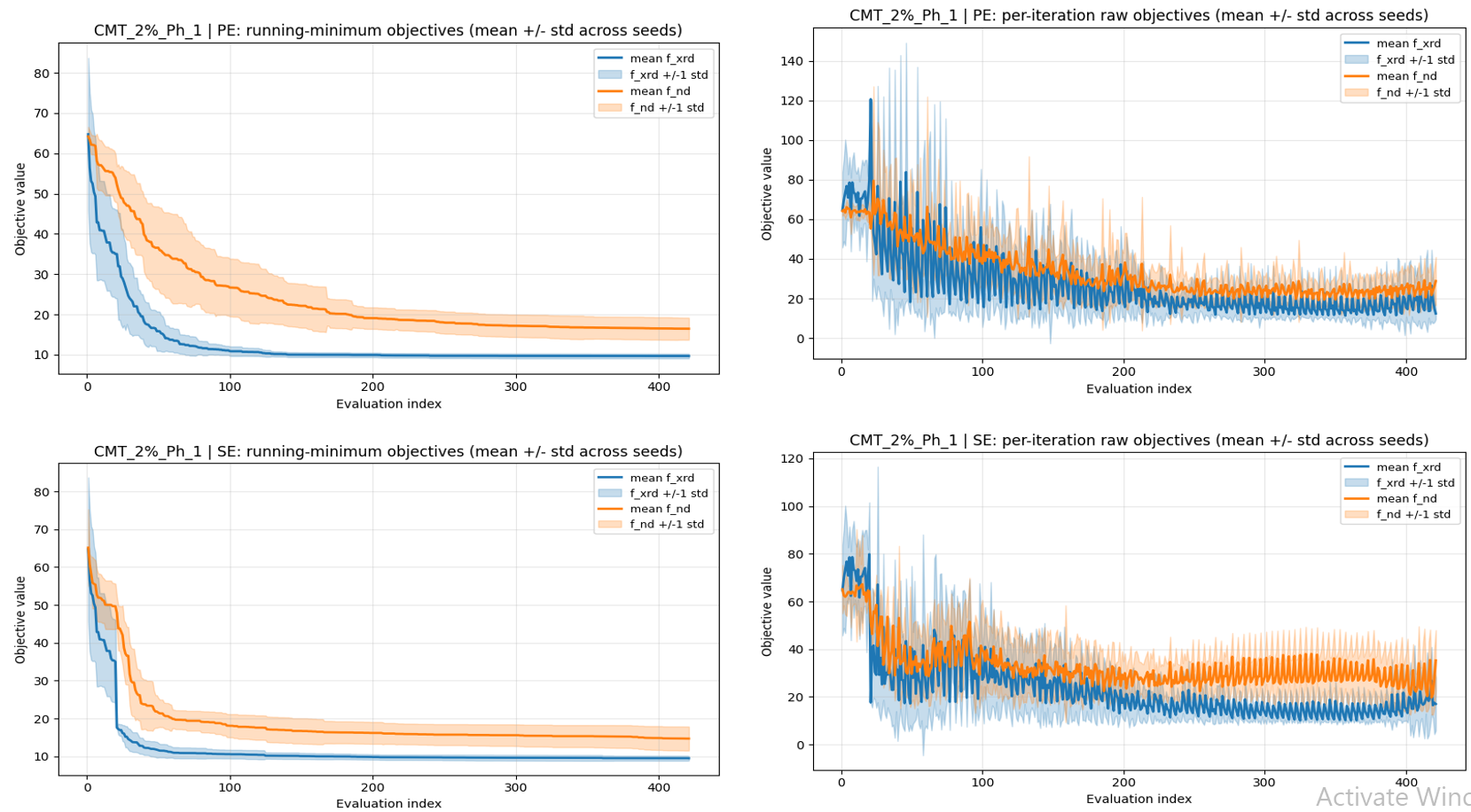}
    \caption{Figures showing convergence curve for the multiple run of the experiments for CMT Pase 1 only PE and SE 2\% bounding box width. Values plotted are means and standard deviations of objective values. Column 1 and 2 for each methods are running minimumns and raw objective values respectively.}
    \label{fig:appendix_five}
    \vspace{-2mm}
\end{figure}

\begin{figure}[t]
    \centering
    \includegraphics[width=\linewidth]{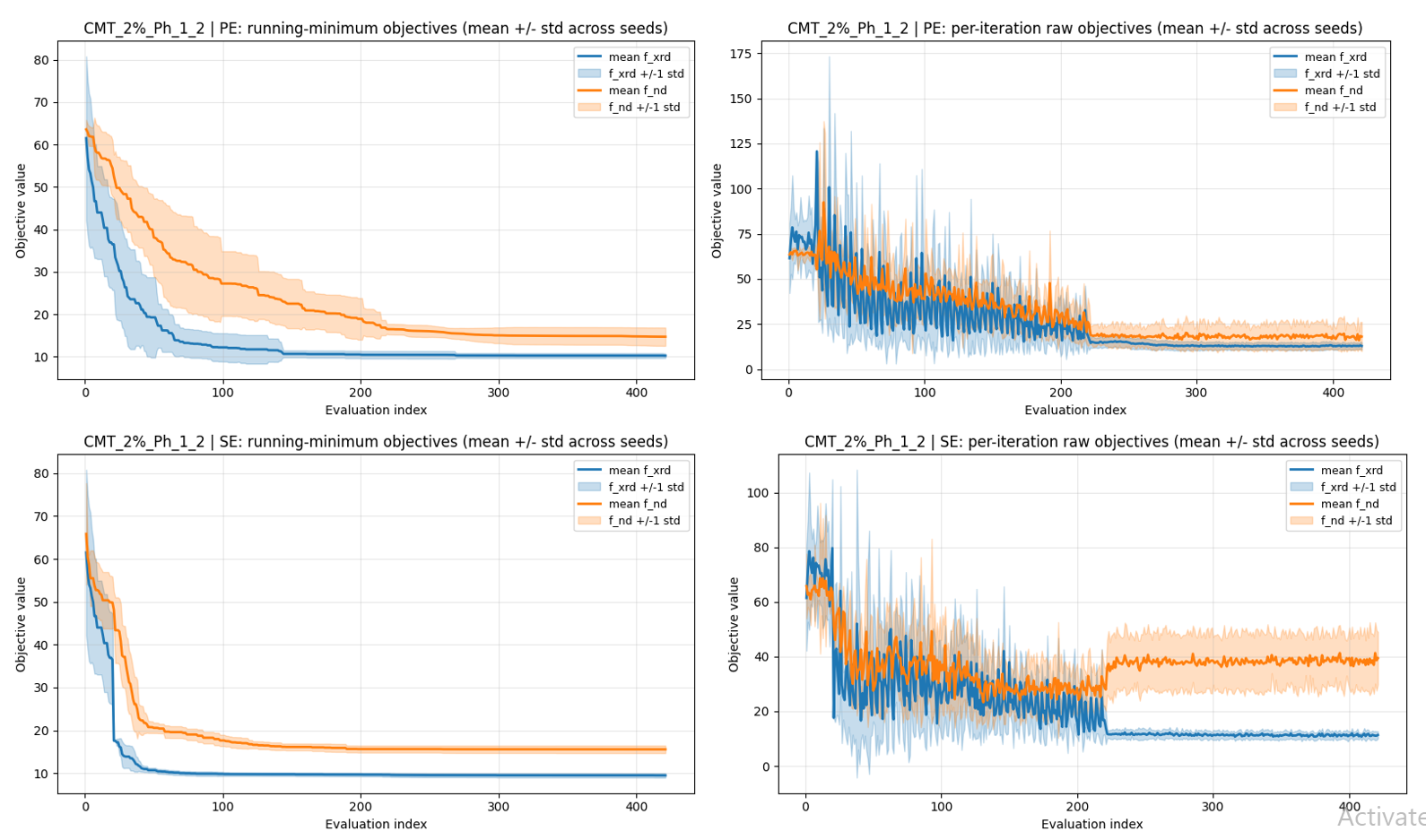}
    \caption{Figures showing convergence curve for the multiple run of the experiments for CMT Pase 1 and 2 PE and SE 2\% bounding box width. Values plotted are means and standard deviations of objective values. Column 1 and 2 for each methods are running minimumns and raw objective values respectively.}
    \label{fig:appendix_six}
    \vspace{-2mm}
\end{figure}

\begin{figure}[t]
    \centering
    \includegraphics[width=\linewidth]{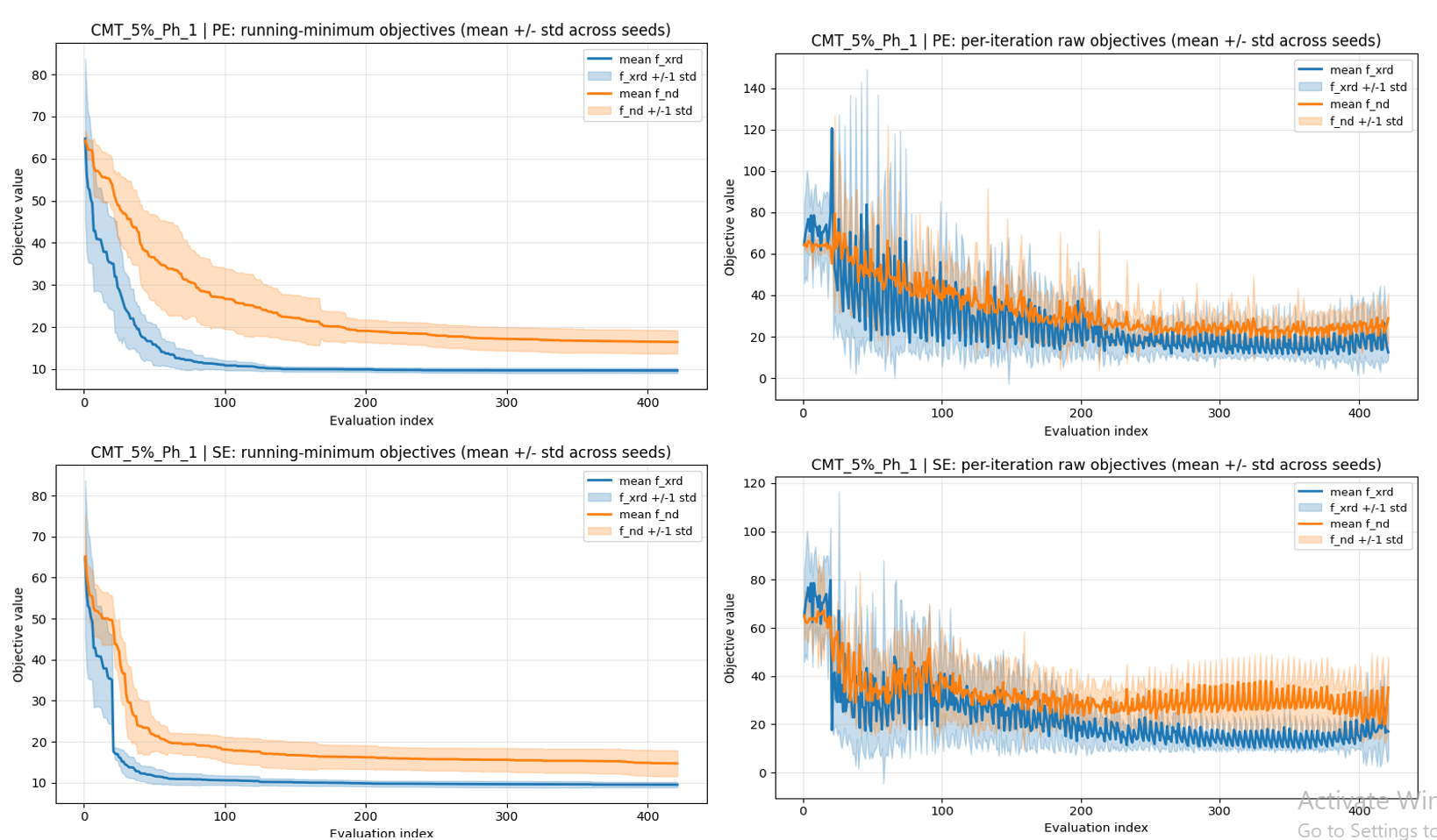}
    \caption{Figures showing convergence curve for the multiple run of the experiments for CMT Pase 1 only PE and SE 5\% bounding box width. Values plotted are means and standard deviations of objective values. Column 1 and 2 for each methods are running minimumns and raw objective values respectively.}
    \label{fig:appendix_seven}
    \vspace{-2mm}
\end{figure}

\begin{figure}[t]
    \centering
    \includegraphics[width=\linewidth]{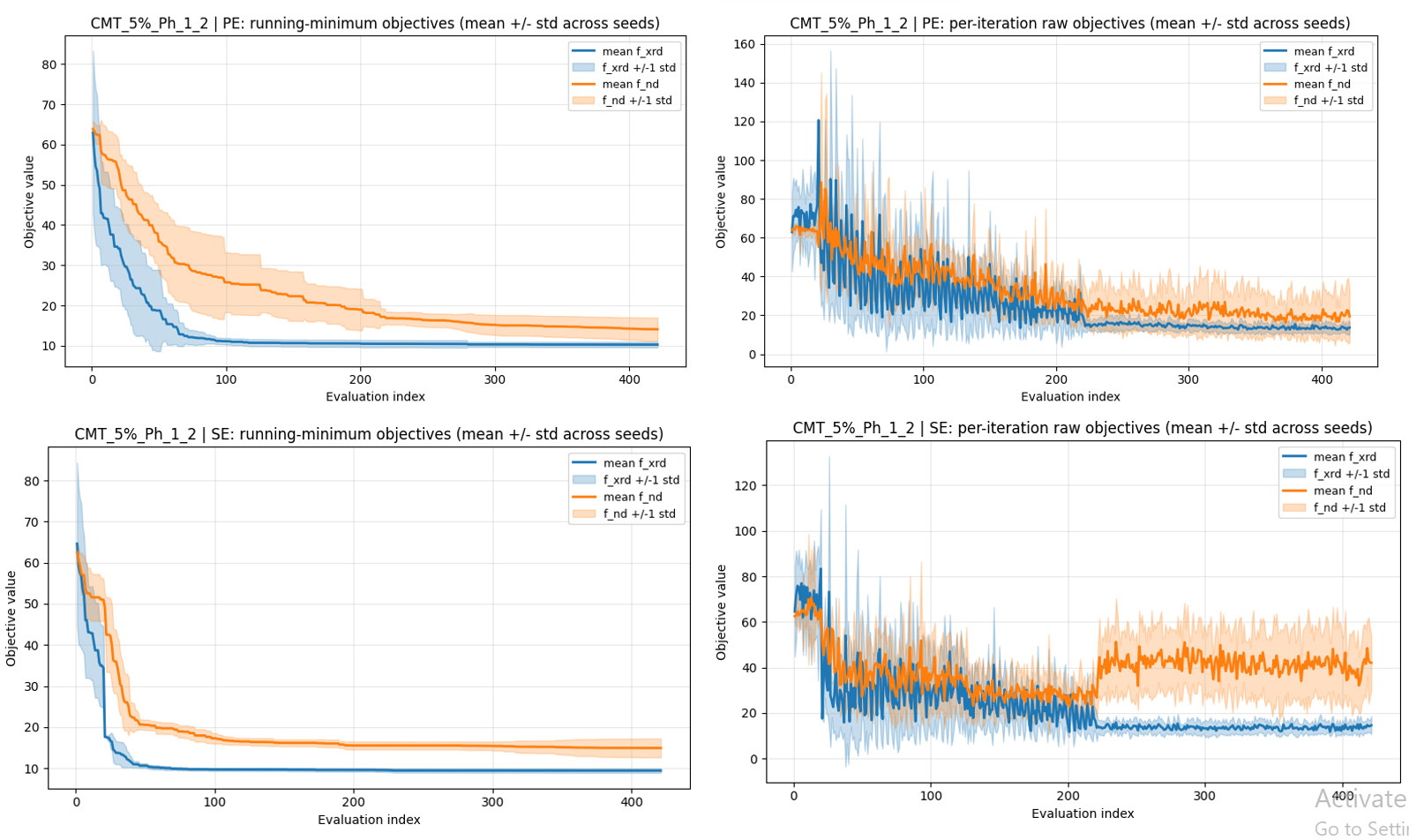}
    \caption{Figures showing convergence curve for the multiple run of the experiments for CMT Pase 1 and 2 PE and SE 5\% bounding box width. Values plotted are means and standard deviations of objective values. Column 1 and 2 for each methods are running minimumns and raw objective values respectively.}
    \label{fig:appendix_eight}
    \vspace{-2mm}
\end{figure}

\begin{figure}[t]
    \centering
    \includegraphics[width=\linewidth]{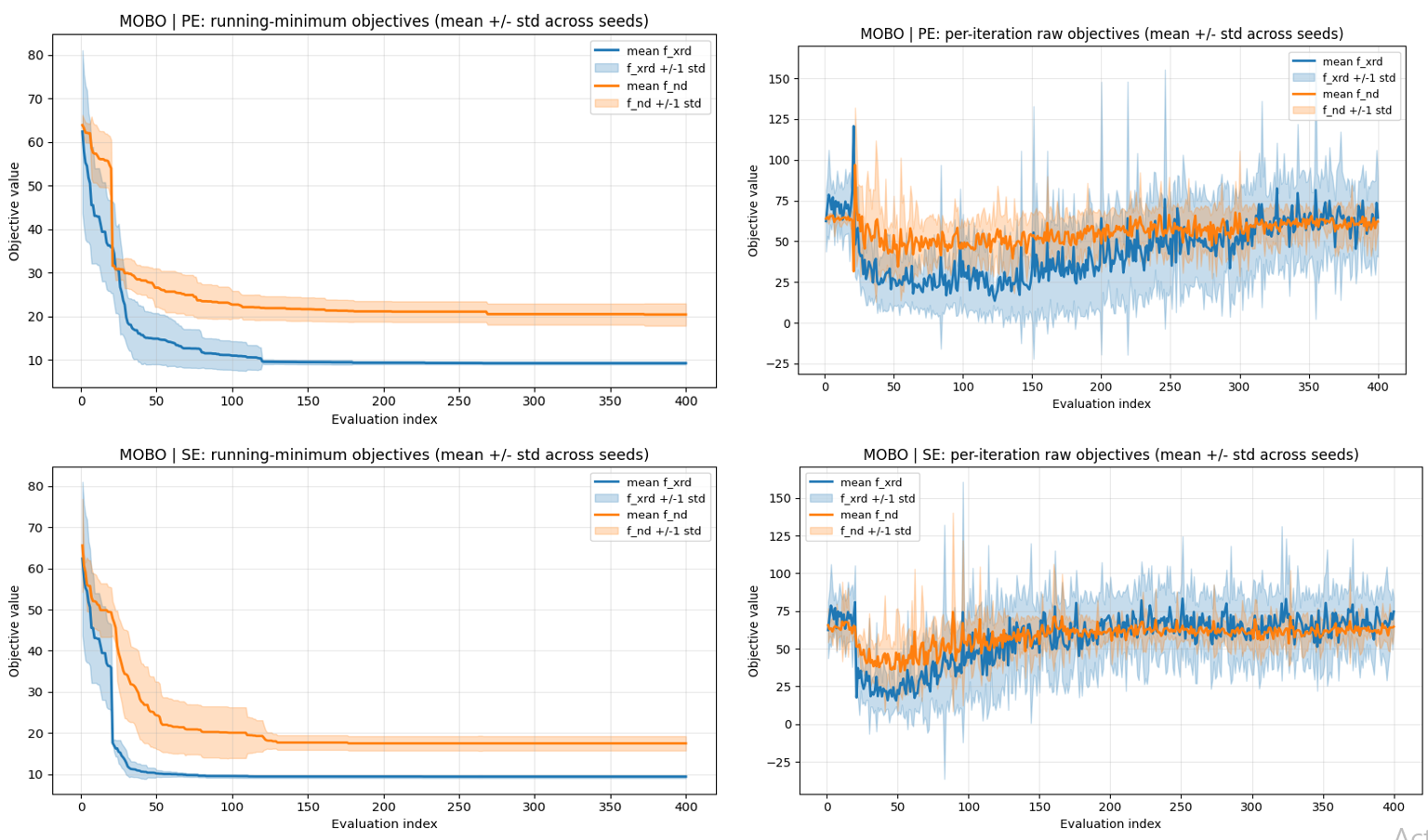}
    \caption{Figures showing convergence curve for the multiple run of the experiments for MOBO. Values plotted are means and standard deviations of objective values. Column 1 and 2 for each methods are running minimumns and raw objective values respectively.}
    \label{fig:appendix_nine}
    \vspace{-2mm}
\end{figure}

\begin{figure}[t]
    \centering
    \includegraphics[width=\linewidth]{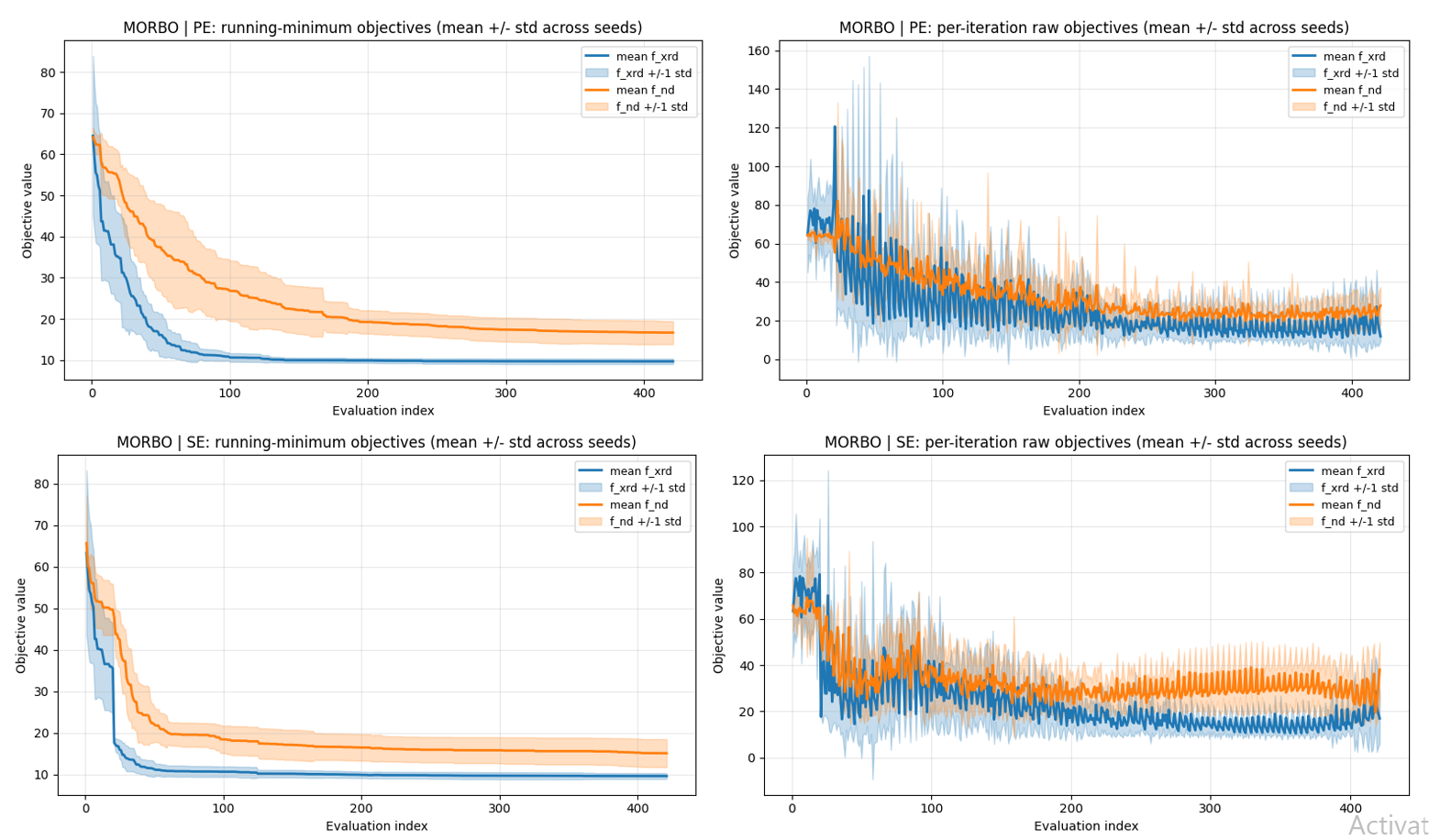}
    \caption{Figures showing convergence curve for the multiple run of the experiments MORBO. Values plotted are means and standard deviations of objective values. Column 1 and 2 for each methods are running minimumns and raw objective values respectively.}
    \label{fig:appendix_ten}
    \vspace{-2mm}
\end{figure}

\begin{figure}[t]
    \centering
    \includegraphics[width=\linewidth]{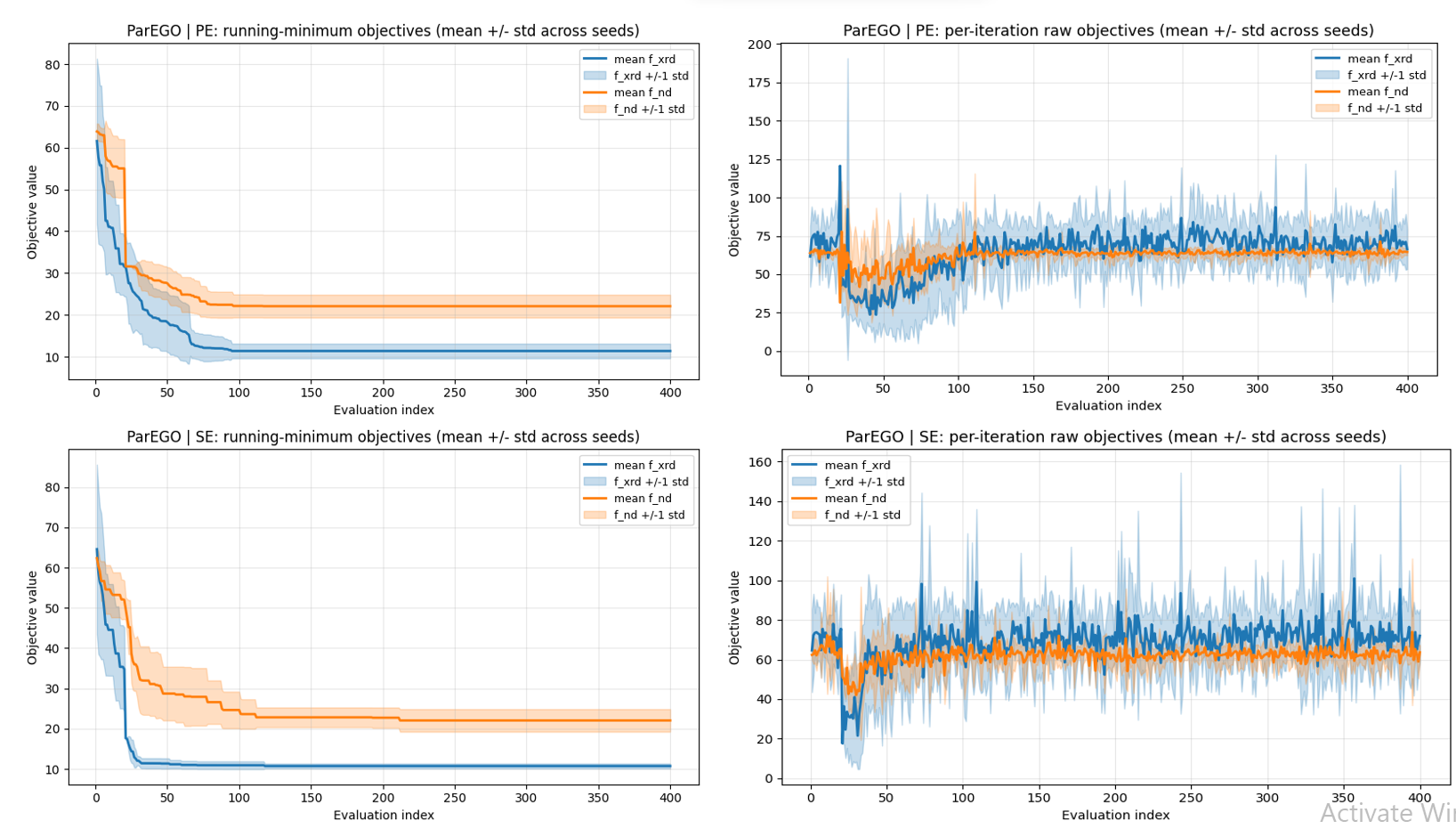}
    \caption{Figures showing convergence curve for the multiple run of the experiments ParEGO. Values plotted are means and standard deviations of objective values. Column 1 and 2 for each methods are running minimumns and raw objective values respectively.}
    \label{fig:appendix_eleven}
    \vspace{-2mm}
\end{figure}

\newpage

\end{document}